\documentclass[sigconf, nonacm]{acmart}

\newcommand\vldbdoi{XX.XX/XXX.XX}
\newcommand\vldbpages{XXX-XXX}
\newcommand\vldbvolume{14}
\newcommand\vldbissue{1}
\newcommand\vldbyear{2021}

\usepackage{xspace}
\newcommand{\sys}{\textsc{GMLP}\xspace}

\usepackage{subfigure}
\usepackage{multirow}
\usepackage{bm}
\usepackage{bbm}
\usepackage{color, soul}
\usepackage{ulem}
\usepackage{subfigure}
\usepackage{graphicx}
\usepackage{listings}
\usepackage{xcolor}

\usepackage{makecell}
\usepackage{color, soul}

\usepackage{bm}
\usepackage{booktabs}

\usepackage[linesnumbered,vlined,ruled,noend]{algorithm2e}

\definecolor{codegray}{rgb}{0.5,0.5,0.5}

\usepackage{enumitem}

\usepackage{wrapfig}

\begin{document}
\title{\sys: Building Scalable and Flexible Graph Neural
Networks with Feature-Message Passing
}

\renewcommand{\shorttitle}{Graph Multi-layer Perceptron}

\author{Wentao Zhang$^{\dagger\ddagger}$, Yu Shen$^\dagger$, Zheyu Lin$^\dagger$, Yang Li$^\dagger$, Xiaosen Li$^\ddagger$, Wen Ouyang$^\ddagger$}
\author{Yangyu Tao$^\ddagger$, Zhi Yang$^\dagger$, Bin Cui$^\dagger$}
\affiliation{
$^\dagger$EECS, Peking University~~~~~$^\ddagger$Data Platform, Tencent Inc.}
\affiliation{
$^\dagger$\{wentao.zhang, shenyu, linzheyu, liyang.cs, yangzhi, bin.cui\}@pku.edu.cn\\ $^\ddagger$\{wtaozhang, hansenli, gdpouyang, brucetao\}@tencent.com
}
\renewcommand{\shortauthors}{Zhang et al.}

\begin{abstract}
In recent studies, neural message passing has proved to be an effective way to design graph neural networks (GNNs), which have achieved state-of-the-art performance in many graph-based tasks. 
However, current neural-message passing architectures typically need to perform an expensive recursive neighborhood expansion in multiple rounds and consequently suffer from a scalability issue. Moreover, most existing neural-message passing schemes are inflexible since they are restricted to fixed-hop neighborhoods and insensitive to the actual demands of different nodes. 
We circumvent these limitations by a novel feature-message passing framework, called Graph Multi-layer Perceptron (\sys), which separates the neural update from the message passing. With such separation, \sys significantly improves the scalability and efficiency by performing the message passing procedure in a pre-compute manner, and is flexible and adaptive in leveraging node feature messages over various levels of localities. 
We further derive novel variants of scalable GNNs under this framework to achieve the best of both worlds in terms of performance and efficiency.
We conduct extensive evaluations on 11 benchmark datasets, including large-scale datasets like ogbn-products and an industrial dataset, demonstrating that GMLP achieves not only the state-of-art performance, but also high training scalability and efficiency.
\end{abstract}

\settopmatter{printfolios=true}
\maketitle

\begingroup\small\noindent\raggedright\textbf{PVLDB Reference Format:}\\
Wentao Zhang, Yu Shen, Zheyu Lin, Yang Li, Xiaosen Li, Wen Ouyang, Yangyu Tao,  Zhi Yang, and Bin Cui. \sys: Building Scalable and Flexible Graph Neural
Networks with Feature-message Passing. PVLDB, \vldbvolume(\vldbissue): \vldbpages, \vldbyear.\\
\href{https://doi.org/\vldbdoi}{doi:\vldbdoi}
\endgroup
\begingroup
\renewcommand\thefootnote{}\footnote{\noindent
This work is licensed under the Creative Commons BY-NC-ND 4.0 International License. Visit \url{https://creativecommons.org/licenses/by-nc-nd/4.0/} to view a copy of this license. For any use beyond those covered by this license, obtain permission by emailing \href{mailto:info@vldb.org}{info@vldb.org}. Copyright is held by the owner/author(s). Publication rights licensed to the VLDB Endowment. \\
\raggedright Proceedings of the VLDB Endowment, Vol. \vldbvolume, No. \vldbissue\ %
ISSN 2150-8097. \\
\href{https://doi.org/\vldbdoi}{doi:\vldbdoi} \\
}\addtocounter{footnote}{-1}\endgroup

\section{Introduction}
Graph neural networks (GNNs)~\cite{wu2020comprehensive} have become the state-of-the-art method in many supervised and semi-supervised graph representation learning scenarios such as node classification~\cite{gao2018large, hamilton2017inductive, DBLP:conf/iclr/KipfW17, klicpera2018predict, monti2017geometric, DBLP:conf/icml/XuLTSKJ18}, link prediction~\cite{zhang2018link, zhang2017weisfeiler, cai2020multi}, recommendation~\cite{berg2017graph, ying2018graph, monti2017geometric}, and knowledge graphs~\cite{wang2018deep, wang2018zero, lee2018multi, marino2016more}. 
The majority of these GNNs can be described in terms of the neural message passing (NMP) framework~\cite{hamilton2017inductive}, which is based on the core idea of recursive neighborhood aggregation. Specifically, during each iteration, the representation of each node is updated (with neural networks) based on messages received from its neighbors. Despite their success, existing GNN algorithms suffer from two major drawbacks:

First, the current NMP procedure needs to repeatedly perform a recursive neighborhood expansion at each training iteration in order to compute the hidden representations of a given node, which are inherently less scalable due to the expensive computation and communication cost, especially for large-scale graphs. 
For example, a recent snapshot of the Tencent Wechat friendship graph is comprised of 1.2 billion nodes and more than 100 billion edges. Also, the high-dimensional features (typically ranging from 300 to 600) are associated with nodes, making GNNs difficult to be trained efficiently under time and resource constraints. 
While there has been an ever-growing interest in graph sampling strategies~\cite{hamilton2017inductive,ying2018graph,DBLP:conf/iclr/ChenMX18,DBLP:conf/iclr/ZengZSKP20} to alleviate the scalability and efficiency issues, sampling strategy may cause information loss. 
Also, our empirical observation shows that sampling cannot effectively reduce the high ratio of communication time over computation time, leading to IO bottleneck.
This requires a new underlying NMP to achieve the best of both worlds.

Moreover, the existing NMP schemes are inflexible since they are restricted to a fixed-hop neighborhood and insensitive to actual demands of messages. 
This either makes that long-range dependencies cannot be fully leveraged due to limited hops/layers, or loses local information due to introducing many irrelevant nodes and unnecessary messages when increasing the number of hops (i.e., over-smoothing issue ~\cite{chen2020measuring,li2018deeper,DBLP:conf/icml/XuLTSKJ18}). Moreover, no correlations are exploited on different hops of locality. 
As a result, most state-of-the-art GNN models are
designed with two layers (i.e., steps).
These issues prevent existing methods from unleashing their full potential in terms of prediction performance.

\begin{figure}
	\centering
	\includegraphics[width=0.97\columnwidth]{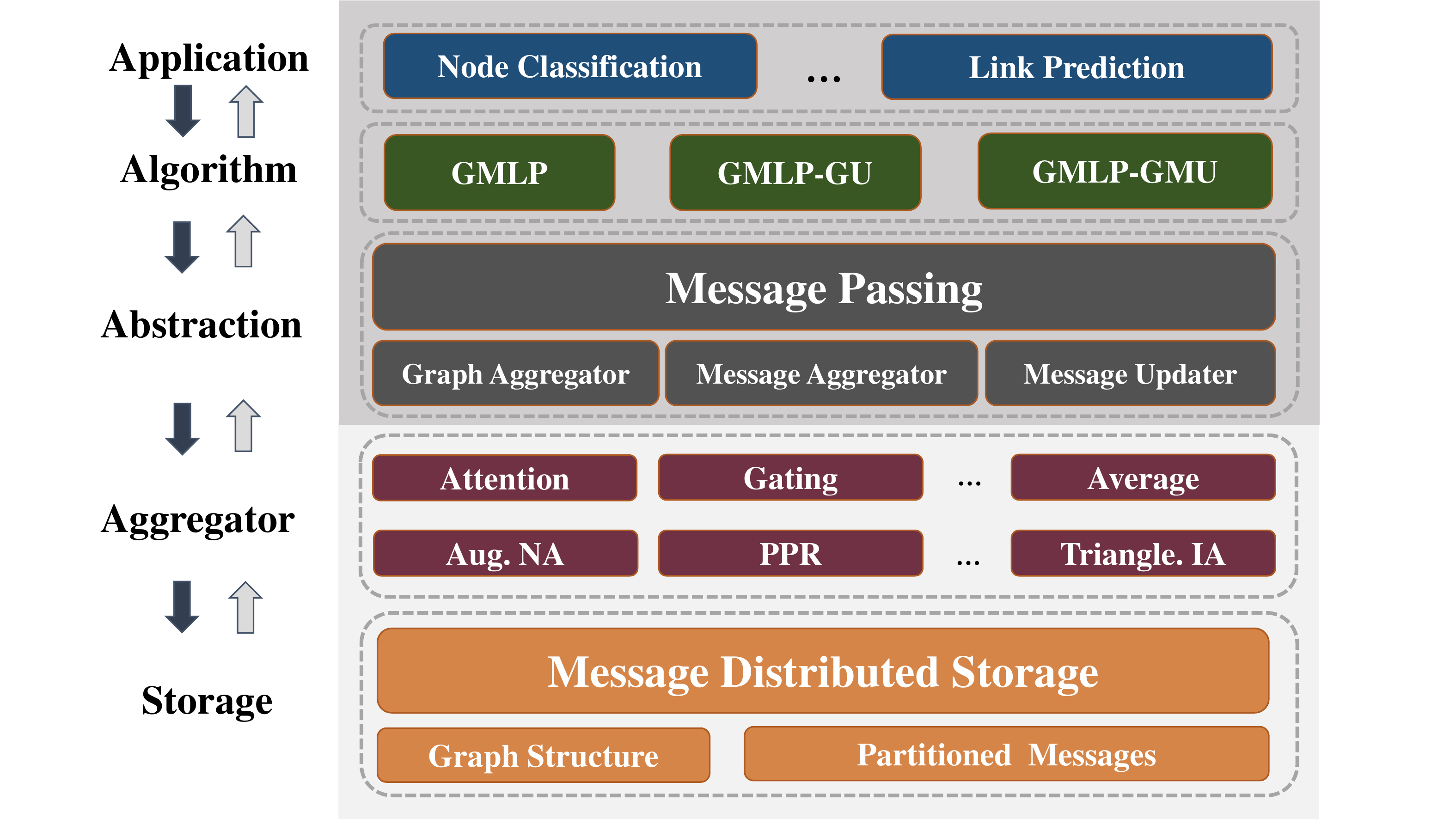}
	\caption{The architecture of \sys framework.}
	\label{fig:arch}
\end{figure}

To address the above limitations, we propose a novel GNN framework called Graph Multi-layer Perceptron (GMLP), which pushes the boundary of the prediction performance towards the direction of \emph{scalable} graph deep learning.
Figure~\ref{fig:arch} illustrates the architecture of our framework. The main contribution of \sys is a new feature message passing (FMP) abstraction, described by graph\_aggregate, message\_aggregate, and update functions. 
This abstraction separates the neural network update from the message passing and leverages multiple messages over different levels of localities to update the node's final representation.
Specifically, instead of messages updated by neural networks, \sys passes node features and avoids repeated expensive message passing procedure by pre-processing it in a distributed manner, thus achieving high scalability and efficiency. Meanwhile, \sys is highly flexible by providing a variety of graph aggregators and message aggregators for generating and combining messages at different localities, respectively. 
Under this new abstraction, we could explore many novel scalable GNN variations within this framework, permitting a more flexible and efficient accuracy-efficiency tradeoff. 
For instance, through adaptively combining the message on multiple localities for each node with a new self-guidance attention scheme, our novel variant could achieve much better accuracy than existing GNNs while maintaining high scalability and efficiency. 
We also find recently emerging scalable algorithms, such as SGC~\cite{wu2019simplifying} and SIGN~\cite{sign_icml_grl2020}, are special variants of our \sys framework.

To validate the effectiveness of \sys, we conduct extensive experiments on 11 benchmark datasets against various GNN baselines. Experimental results demonstrate that \sys outperforms the state-of-the-art methods APPNP and GAT by a margin of 0.2-3.0\% and 0.2\%-3.6\% in terms of predictive accuracy, while achieving up to $15.6\times$ and $74.4\times$ training speedups, respectively. Moreover, \sys achieves a nearly linear speedup in distributed training.

Our contributions can be summarized as follows:
    {\it (1) Abstraction.} We present a novel message passing abstraction that supports efficient and scalable feature aggregation over node neighborhoods in an adaptive manner.
    {\it (2) Algorithms.} We explore various scalable and efficient \sys variants under the new abstraction with carefully designed message passing and aggregation strategies.  
    {\it (3) Performance and efficiency.} The experimental results on massive datasets demonstrate that \sys framework outperforms the existing GNN methods and meanwhile achieves better efficiency and scalability.
    \vspace{-0.8em}
\section{Preliminary}
In this section, we introduce GNNs from the view of message passing (MP) along with the corresponding scalability challenge. 

\subsection{GNNs and Message-Passing}
\label{mp}
Considering a graph $\mathcal{G(V, E)}$ with nodes $\mathcal{V}$, edges $\mathcal{E}$ and features for all nodes $\mathbf{x}_v \in \mathbb{R}^d, \forall v \in \mathcal{V}$, many proposed GNN models can be analyzed using the message passing (MP) framework.

\paragraph{\emph{Neural message passing (NMP).}} This NMP framework is widely adopted by mainstream GNNs, like GCN~\cite{DBLP:conf/iclr/KipfW17}, GraphSAGE~\cite{hamilton2017inductive}, GAT~\cite{DBLP:conf/iclr/VelickovicCCRLB18}, and GraphSAINT~\cite{DBLP:conf/iclr/ZengZSKP20}, where each layer adopts a neighborhood and an updating function.
At timestep $t$, a message vector $\mathbf{m}^{t}_v$ for node $v\in \mathcal{V}$ is computed with the representations of its neighbors $\mathcal{N}_v$ using an aggregate function, and $\mathbf{m}^{t}_v$ is then updated by a neural-network based update function, which is:
\begin{equation}
\begin{aligned}
& \mathbf{m}^t_v \gets \texttt{aggregate}\left(\left\{\mathbf{h}^{t-1}_u|{u \in \mathcal{N}_v}\right\}\right),\ \mathbf{h}^t_v \gets \texttt{update}(\mathbf{m}^t_v).
\end{aligned}
\label{eq:mp}
\end{equation}
Messages are passed for $T$ timesteps so that the steps of message passing correspond to the network depth. Taking the vanilla GCN~\cite{DBLP:conf/iclr/KipfW17} as an example, we have:
\begin{equation}
\begin{aligned}
 &\texttt{GCN-aggregate}\left(\left\{\mathbf{h}^{t-1}_u|{u \in \mathcal{N}_v}\right\}\right)=\sum_{u \in \mathcal{N}_v}\mathbf{h}_u^{t-1}/\sqrt{\tilde{d}_v\tilde{d}_u},\\
  &\texttt{GCN-update}(\mathbf{m}^t_v)=\sigma(W\mathbf{m}^t_v),\nonumber
 \end{aligned}
\end{equation}
where $\tilde{d}_v$ is the degree of node $v$ obtained from the adjacency matrix with self-connections $\tilde{A}=I+A$. 

\paragraph{\emph{Decoupled neural message passing (DNMP)}} Note that the aggregate and update operations are inherently intertwined in equation~\eqref{eq:mp}, i.e., each aggregate operation requires a neural layer to update the node‘s hidden state in order to generate a new message for the next step. Recently, some researches show that such entanglement could compromise performance on a range of benchmark tasks~\cite{wu2019simplifying, sign_icml_grl2020}, and suggest separating GCN from the aggregation scheme.
We reformulate these models into a single decoupled neural MP framework: Neural prediction messages are first generated (with update function) for each node
utilizing only that node’s own features, and then aggregated using aggregate function.
\begin{equation}
\begin{aligned}
& \mathbf{h}^{0}_v \gets \texttt{update}(\mathbf{x_v}),\  \mathbf{h}^t_v \gets \texttt{aggregate}\left(\left\{\mathbf{h}^{t-1}_u|{u \in \mathcal{N}_v}\right\}\right).
\end{aligned}
\label{eq:dmp}
\end{equation}
where $x_v$ is the input feature of node $v$. 
Existing methods, such as PPNP~\cite{klicpera2018predict}, APPNP~\cite{klicpera2018predict}, AP-GCN~\cite{spinelli2020adaptive} and etc., follows this decoupled MP. Taking APPNP as an example:
\begin{equation}
\begin{aligned}
&\texttt{APPNP-update}(\mathbf{x_v})=\sigma(W\mathbf{x}_v),\\
&\texttt{APPNP-aggregate}\left(\left\{\mathbf{h}^{t-1}_u|{u \in \mathcal{N}_v}\right\}\right)=\alpha \mathbf{h}_v^0+(1-\alpha)\sum_{u \in \mathcal{N}_v}\frac{\mathbf{h}_u^{t-1}}{\sqrt{\tilde{d}_v\tilde{d}_u}}, 
\end{aligned}\nonumber
\end{equation}
where \texttt{aggregate} function adopts personalized PageRank with the restart probability $\alpha \in \left(0,1 \right]$ controlling the locality.

\subsection{Challenges and Motivation}\label{sec:observation}
\paragraph{\emph{Scalable training}}Most researches on GNNs are tested on small benchmark graphs, while in practice, graphs contain billions of nodes and edges. 
However, we find that neural message passing formulated in equations~\eqref{eq:mp} and \eqref{eq:dmp} do not scale well to large graphs since they typically need to repeatedly perform a recursive neighborhood expansion to gather \emph{neural messages}. 
In other words, both MP and decoupled MP generate new messages (e.g., $\mathbf{h_v}, \forall v\in \mathcal{V}$) during each training epoch, and the expensive aggregate procedure need to be performed again, which requires gathering from its neighbors, and the neighbors in turn, have to gather from their own neighbors and so on. This process leads to a costly recursive neighborhood expansion growing with the number of layers.
For large graphs that can not be entirely stored on each worker's local storage~\cite{ma2019neugraph, zhang2020agl, zhu2019aligraph}, gathering the required neighborhood neural messages leads to massive data communication costs~\cite{ tripathy2020reducing, zheng2020distdgl, lin2020pagraph}.

\begin{figure}
 \centering
 \subfigure[Speedup]{
  \scalebox{0.45}[0.45]{
   \includegraphics[width=1\linewidth]{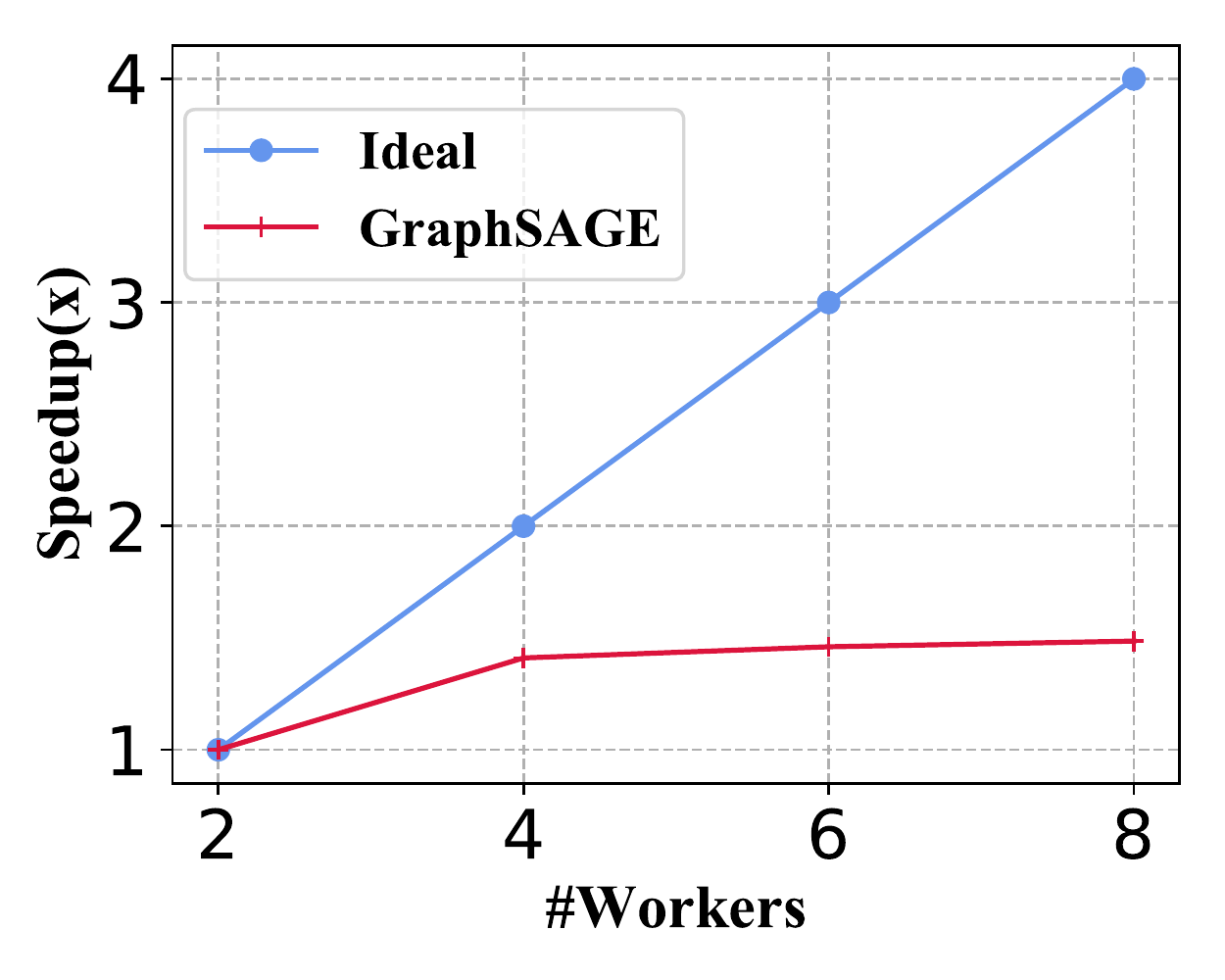}
 }}
 \subfigure[Bottleneck]{
  \scalebox{0.45}[0.45]{
   \includegraphics[width=1\linewidth]{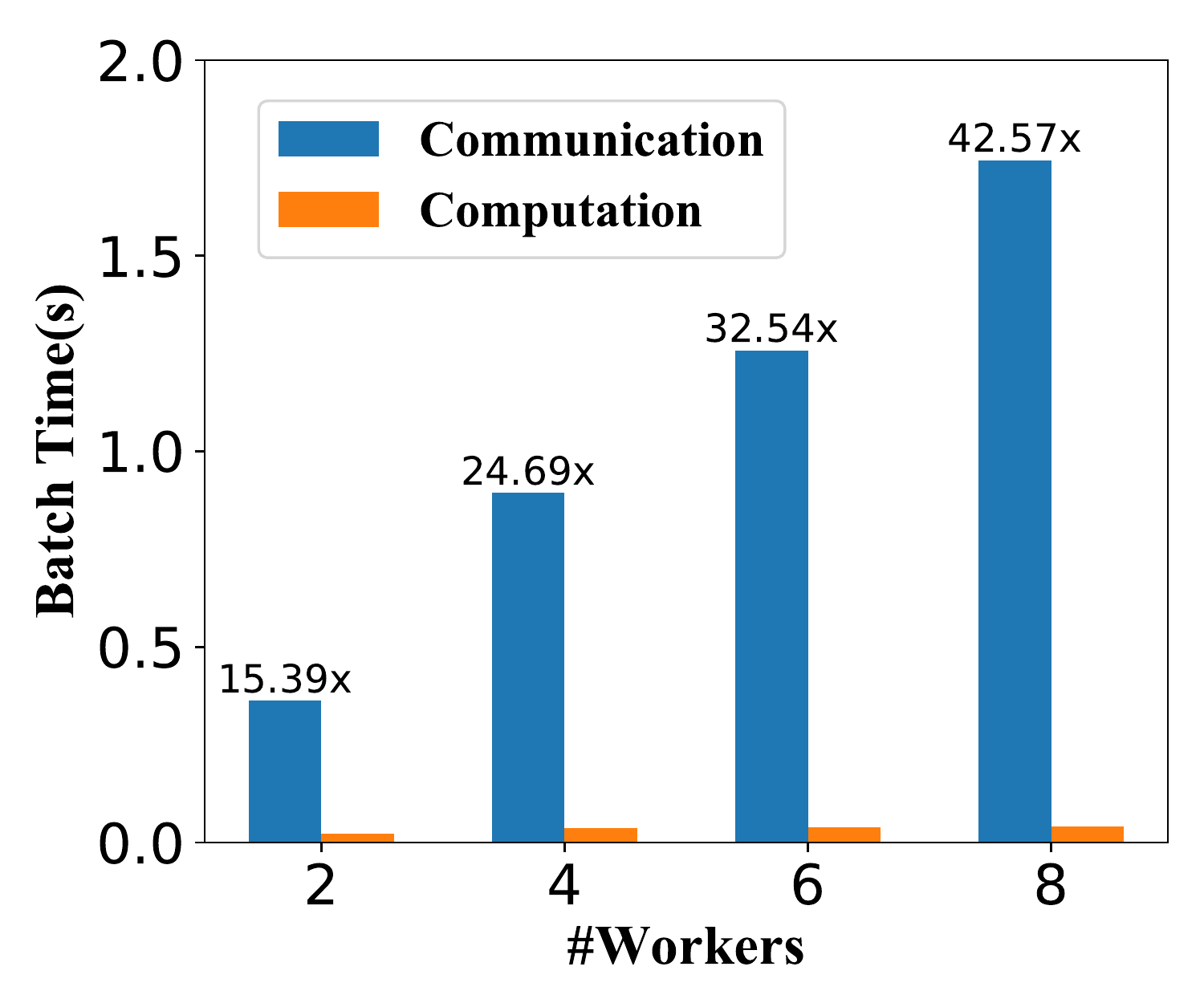}
 }}
\vspace{-4mm}
\caption{The speedup and bottleneck of a two-layer GraphSAGE along with the increased workers on Reddit dataset.}
\label{fig:unscale}
\end{figure}

To demonstrate the scalability issue, we utilize distributed training functions provided by DGL~\cite{bronstein2017geometric} to test the scalability of GraphSAGE~\cite{hamilton2017inductive} (batch size = 8192). We partition the Reddit dataset across multiple machines, and treat each GPU as a worker.
Figure ~\ref{fig:unscale} illustrates the training speedup along with the number of workers and the bottleneck in distributed settings. In particular, Figure ~\ref{fig:unscale}(a) shows that the scalability of GraphSAGE is significantly limited even when the mini-batch training and graph sampling method are adopted. Note that the speedup is calculated relative to the runtime of two workers.
Figure (b) further demonstrates that the scalability is mainly bottlenecked by the aggregation procedure in which high data loading cost is required to gather multi-scale neural messages.
Other GNNs under the NMP (or DNMP) framework suffer from the same scalability issue as they perform such costly aggregation during each training epoch. This motivates us to separate the neural update and message passing towards scalable training.

\begin{figure}
    \centering
    \includegraphics[width=\linewidth]{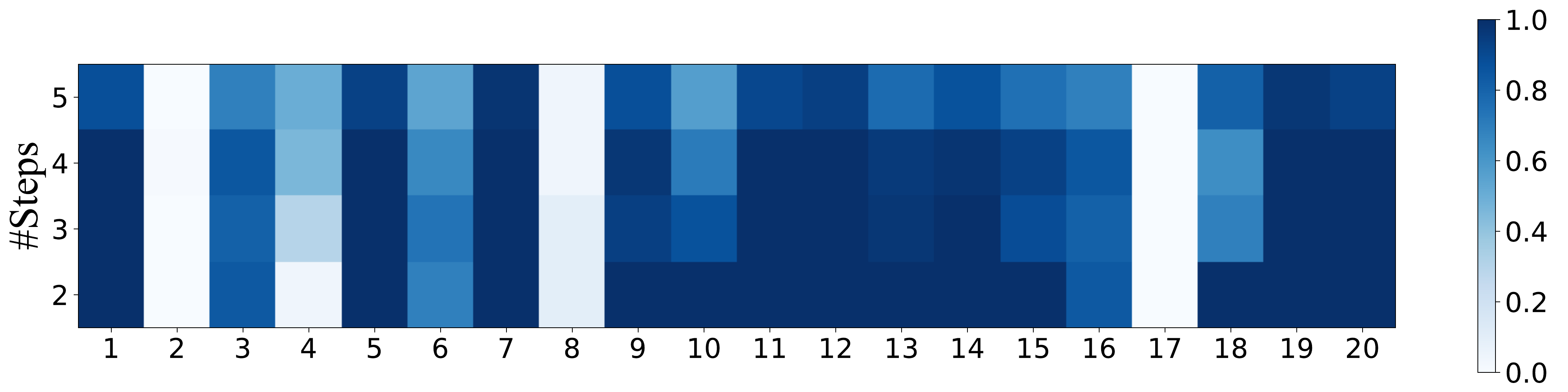}
    \vspace{-7mm}
    \caption{The influence of aggregation steps on 20 randomly sampled nodes on Citeseer dataset. The X-axis is the node id and Y-axis is the aggregation steps (number of layers in GCN). The color from white to blue represents the ratio of being predicted correctly in 50 different runs.}
    \label{fig:flexibility}
\end{figure}

\vspace{-0.8em}
\paragraph{\emph{Model Flexibility.}}
We find that both NMP and DNMP is inflexible given a fixed $T$ step of aggregation for all nodes. It is difficult to determine a suitable $T$. Few steps may fail to capture sufficient neighborhood information, while more steps may bring too much information which leads to the over-smoothing issue. Moreover, nodes with different neighborhood structure require different message passing steps to fully capture the structural information.
As shown in Figure \ref{fig:flexibility}, we apply standard GCN with different layers to conduct node classification on the Cora dataset. 
We observe that most of the nodes are well classified with two steps, and as a result, most state-of-the-art GNN models are designed with two layers (i.e., steps). In addition, the predictive accuracy on 13 of the 20 sampled nodes increases with a certain step larger than two.
The above observation motivates us to design a flexible and adaptive aggregate function for each node. However, this becomes even more challenging when complex graphs and sparse features are given.
\section{GMLP Abstraction}
To address the above challenges, we propose a new GMLP abstraction under which more flexible and scalable GNNs can be derived. 
\subsection{Feature-message Passing Interfaces}
GMLP consists of a message aggregation phase over graph followed by message combination and applying phases. 

\paragraph{\underline{Interfaces}}
At timestep $t$, a message vector $\mathbf{m}^{t}_v$ is collected from the messages of the neighbors $\mathcal{N}_v$ of $v$:
\begin{equation}
\mathbf{m}^t_v \gets \texttt{graph\_aggregator}\left(\left\{\mathbf{m}^{t-1}_u|{u \in \mathcal{N}_v}\right\}\right),
\end{equation}
where $\mathbf{m}^0_v=\mathbf{x}_v$. The messages could be passed for $T$ timesteps, where $m^t_v$ at timestep $t$ could gather the neighborhood information from nodes that are $t$-hop away. 
The multi-hop messages $\mathcal{M}_v=\{\mathbf{m}^{t}_v\ |\ 0\leq t \leq T\}$ are then aggregated into a single combined message vector $c_v$ for a node $v$ such that the model learns to combine different scales for a given node:
\begin{equation}
\mathbf{c}^T_v \gets \texttt{message\_aggregator}(\mathcal{M}_v, \mathbf{r}_v),
\end{equation}
where $\mathbf{r}_v$ is a reference vector for adaptive message aggregation on node $v$ (null by default).
The GMLP learns node representation $\mathbf{h}_v$ by applying an MLP network to the combined message vector $\mathbf{c}^T_v$:
\begin{equation}
\mathbf{h}_v \gets \texttt{udpate}(\mathbf{c}_v).
\end{equation}

\begin{figure} 
	\centering
	\begin{minipage}[t]{0.98\linewidth}
	\begin{lstlisting}
interface GMLP{
  //neighborhood message aggregation 
  msg = graph_aggregator(Nbr_messages);
  //node's message aggregation
  c_msg = message_aggregator(msgList, reference);
  //apply the combined message with MLP
  h = update(c_msg)
};
 	\end{lstlisting} \vspace{-1ex}
	\end{minipage}
	\caption{The GMLP Abstraction.}
	\label{fig:op_interface}
\end{figure}

There are two key differences of the \sys abstraction from existing GNN message passing (MP). 

\paragraph{\underline{Message type}}To address the scalability challenge, the message type in our abstraction is different from the previous GNN MP. We propose to pass node \emph{feature messages} instead of neural messages by making the message aggregation independent of the update of hidden state. 
In particular, MP in the existing GNNs needs to update the hidden state $\mathbf{h}^{t}_v$ by applying the message vector $\mathbf{m}^{t}_v$ with neural networks, in order to perform the aggregate function for next step.
Decoupled-GNN MP also needs to first update the hidden state $\mathbf{h}^{t}_v$ with neural networks in order to get the neural message for the following multi-step aggregate procedure.
By contrast, \sys allows passing the node feature message without applying messages on hidden state in \texttt{graph\_aggregator}. 
This message passing procedure is independent of learnable model parameters and can be easily pre-computed, thus leading to high scalability and speedup.

\paragraph{\underline{Multi-scale Messages}}
The existing MP framework only utilizes the last message vector $\mathbf{m}^{T}_v$ to compute the final hidden state $\mathbf{h}^{T}_v$. Motivated by the observation in Section~\ref{sec:observation}, GMLP assumes that the optimal neighborhood expansion size should not be the same for each node $v$, and thus retaining all the messages $\{\mathbf{m}^t_v|t\in[1,\ T]\}$ that a node $v$ receives over different steps (i.e., localities). 
The multi-scale messages are then aggregated per node into a single vector with \texttt{message\_aggregator}, such that we could balance the preservation of information from both local and extended (multi-hop) neighborhood for a given node. 

\subsection{Graph Aggregators}\label{sec:ga}
To capture the information of nodes that are several hops away, \sys adopts a graph aggregator to combine the nodes with their neighbors during each timestep. 
Intuitively, it is unsuitable to use a fixed graph aggregator for each task since the choice of graph aggregators depends on the graph structure and features. Thus \sys provides three different graph aggregators to cope with different scenarios, and one could add more aggregators following the semantic of \texttt{graph\_aggregator} interface.

\paragraph{\underline{Normalized adjacency}} 
The augmented normalized adjacency (Aug. NA)~\cite{DBLP:conf/iclr/KipfW17} and random walk are simple yet effective on a range of GNNs. The only difference between the two aggregators lies in its normalization method. The former uses the renormalization trick while the latter applies the random walk normalization. We denote $\tilde{d}_v$ as the degree of node $v$ obtained from the augmented adjacency matrix $\tilde{A}=I+A$, the normalized graph aggregator is: 
\begin{small}
\begin{equation}
\mathbf{m}_v^{t}=\sum_{u \in \mathcal{N}_v}\frac{1}{\sqrt{\tilde{d}_v\tilde{d}_u}}\mathbf{m}_u^{t-1}.
\end{equation}
\end{small}

\paragraph{\underline{Personalized PageRank}} 
Personalized PageRank (PPR) ~\cite{DBLP:journals/corr/abs-1810-05997} focuses on its local neighborhood using a restart probability $\alpha \in \left(0,1 \right]$ and performs well on graphs with noisy connectivity. While the calculation of the fully personalized PageRank matrix is computationally expensive, we apply its approximate computation ~\cite{DBLP:journals/corr/abs-1810-05997}:

\begin{small}
\begin{equation}
    \mathbf{m}_v^{t}=\alpha \mathbf{m}_v^0+(1-\alpha)\sum_{u \in \mathcal{N}_v}\frac{1}{\sqrt{\tilde{d}_v\tilde{d}_u}}\mathbf{m}_u^{t-1},
\end{equation}
\end{small}
where the restart probability $\alpha$ allows to balance preserving locality (i.e., staying close to the root node to avoid over-smoothing) and leveraging the information from a large neighborhood.

\paragraph{\underline{Triangle-induced adjacency}} 
Triangle-induced adjacency (Triangle. IA) matrix ~\cite{DBLP:journals/corr/abs-1802-01572} accounts for the higher order structures and helps distinguish strong and weak ties on complex graphs like social graphs. We assign each edge a weight representing the number of different triangles it belongs to, which forms a weight metrix $A^{tri}$. We denote $d^{tri}_v$ as the degree of node $v$ from the weighted adjacency matrix $A^{tri}$. The aggregator is then calculated by applying a row-wise normalization:
\begin{small}
\begin{equation}
    \mathbf{m}_v^{t}=\sum_{u \in \mathcal{N}_v}\frac{1}{d_v^{tri}}\mathbf{m}_u^{t-1}.
\end{equation}
\end{small}

\subsection{Message Aggregators}
Before updating the hidden state of each node, \sys proposes to apply a message aggregator to combine messages obtained by graph aggregators per node into a single vector, such that the subsequent model learns from the multi-scale neighborhood of a given node. 
We summarize the message aggregators \sys supports, and one can also include more aggregators with the future state-of-the-arts.

\paragraph{\underline{Non-adpative aggregators}} 
The main characteristic of these aggregators is that they do not consider the correlation between messages and the center node. The messages are directly concatenated or summed to obtain the combined message vector. 
\begin{equation}
    c_{msg} \gets \oplus_{m_v^i\in M_v}f(m_v^i),
\end{equation}
where $f$ can be a function used to reduce the dimension of message vectors, and the $\oplus$ can be concatenating or pooling operators including average pooling or max pooling.
The $\max$ used in the max-pooling aggregator is an element-wise operator. 
Compared with the pooling operators, although the concatenating operator keeps all the input message information, the dimension of its outputs increases as $T$ grows, leading to additional computational costs on the following MLP network.

\paragraph{\underline{Adaptive aggregators}} 
The observations in Section~\ref{sec:observation} imply that messages of different hops make different contributions to the final performance. This motivates the design of adaptive-aggregation functions, which determines the importance of a node’s message at different ranges rather than fixing the same weights for all nodes. To this end, we propose attention and gating, which generate retainment scores, indicating how much the corresponding messages should be retained in the final combined message. 

As we shall describe in Sec.~\ref{sec:alg}, the \textbf{attention} aggregator takes as input a self-guided reference vector $\mathbf{r}_v$ which captures the personalized property of the each target node $v$:
\begin{small}
\begin{equation}
\begin{aligned}
      & \mathbf{c_{msg}} \gets \sum_{\mathbf{m}_v^i\in M_v} w_i \mathbf{m}_v^i,\ w_i = \frac{\exp( \varphi(\mathbf{m}_v^i, \mathbf{r}_v))}{\sum_{\mathbf{m}_v^i\in M_v}\exp( \varphi(\mathbf{m}_v^i, \mathbf{r}_v))},\ \\
\end{aligned}
\end{equation}
\end{small}
where $\varphi(\mathbf{m}_v^i, \mathbf{r}_v) = \tanh(\bm{W}_1 \mathbf{m}_v^i+\bm{W}_2\mathbf{r}_v)$.

Like attention-based aggregator, the \textbf{gating} aggregator also aggregates the messages adaptively. The main difference is that it adopts a trainable global reference vector that are uniform for all nodes. The formula of the gating aggregator is given as follows,
\begin{small}
\begin{equation}
\begin{aligned}
      & \mathbf{c_{msg}} \gets \sum_{\mathbf{m}_v^i\in M_v} w_i \mathbf{m}_v^i,\  w_i = \sigma(\mathbf{s} \mathbf{m}_v^i),\label{eq:gate}
\end{aligned}
\end{equation}
\end{small}
where $\mathbf{s}$ is a trainable vector shared by all nodes to generate gating scores, and $\sigma$ denotes the sigmoid function.

By utilizing these adaptive message aggregators, \sys is capable of balancing the messages from the multi-scale neighborhood for each node, at the expense of training attention parameters. Given various aggregators, \sys provides various algorithms in Sec.~\ref{sec:alg} balancing high flexible performance-efficiency tradeoff. 

\section{GMLP Algorithms}
\label{sec:alg}
Based on the above message passing abstraction, we propose the following
GMLP algorithms.

\subsection{Algorithm Framework}
\paragraph{\underline{Self-guided \sys}}As shown in Algorithm~\ref{alg:gmlp}, the sequence of operations is ``\texttt{graph\_aggregator} $\rightarrow$ \texttt{message\_aggregator} $\rightarrow$\texttt{update}$\rightarrow$\texttt{message\_aggregator}$\rightarrow$\texttt{update}''. 
Specifically, we first apply a $T$-steps \texttt{graph\_aggregator} to derive the messages over different hops of nodes (Line 2-5). Both long-range dependencies and local information are captured by multiple messages. 
The messages are aggregated per node into a single message vector and the vector is then fed to \texttt{update} (e.g., MLP) to obtain the hidden state vector $\widetilde{h_v}$ (Line 7-8).
Note that $\widetilde{h_v}$ is obtained by aggregating messages uniformly over the fixed aggregation steps for each node.
However, the most suitable aggregation steps for each node should be different. As $\widetilde{h_v}$ captures the relation of different messages of a node $v$ and model performance, we include an adaptive, self-guided adjustment by introducing $\widetilde{h_v}$ as a reference vector of $v$ (Line 9-10) to generate retainment scores. 
These scores are used for messages that carry information from a range of neighborhoods.
These retainment scores measure how much information derived by different propagation layers should be retained to generate the final combined message for each node. 

\begin{algorithm}[t]
  \SetAlgoLined
  \KwIn{Graph $G$, propagation number $K$, feature matrix $\mathbf{X}$.}
  \SetAlgoLined
  \caption{GMLP Framework}
  \label{alg:gmlp}
   $M =\{\mathbf{X}\}$\;
  \For{$1\leq t\leq T$}
  {
    \For{$v\in V$}
    {
        $m^t_v \gets$ \texttt{graph\_aggregator}$(m^{t-1}_{\mathcal{N}_v})$\;
        $M_v = M_v \cup \{m^t_v\}$\;
    }
    }
    \For{$v\in V$}
    {
    $\widetilde{c_v} \gets $\texttt{message\_aggregator}$(M_v,\ null)$\; 
    $\widetilde{h_v} \gets $\texttt{update}$(\widetilde{c_v})$\; 
    $c_v \gets $\texttt{message\_aggregator}$(M_v,\ \widetilde{h_v})$\;
    $h_v \gets $\texttt{update}$(c_v)$\; 
  }
\end{algorithm}

\begin{figure}[t]
    \vspace{-4mm}
	\centering
	\includegraphics[width=0.97\columnwidth]{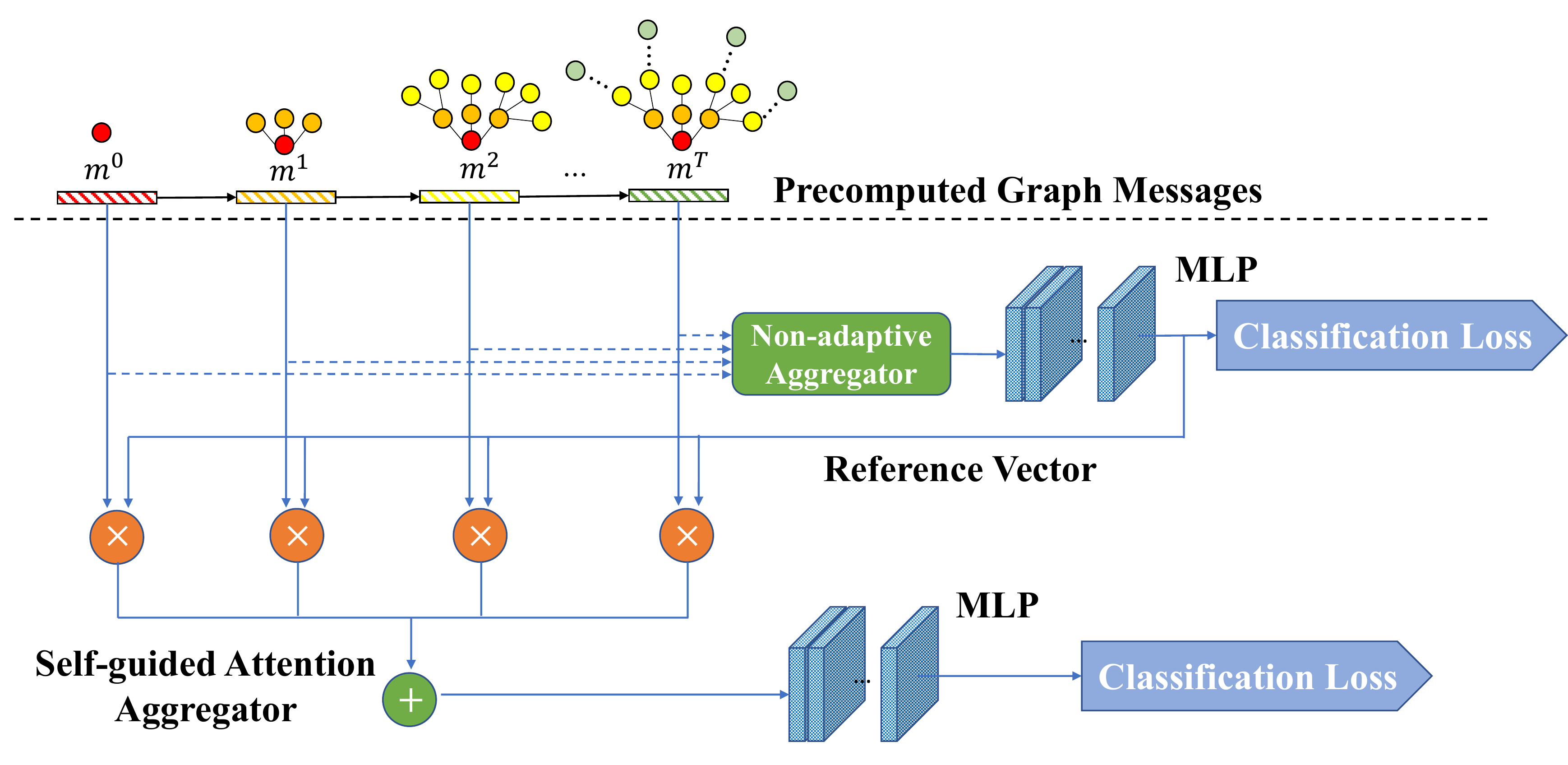}
	\caption{The self-guided network architecture of GMLP.}
	\label{fig:gmlp}
\end{figure}

Figure \ref{fig:gmlp} shows the corresponding model architecture of Algorithm~\ref{alg:gmlp}, which includes two branches: the non-adaptive aggregation (NA) branch (corresponding to Line 7-8 in the algorithm) and self-guided attention aggregation (SGA) branch (corresponding to Line 9-10), which share the same multi-scale graph messages. 
The NA branch aims to create a multi-scale semantic representation of the target node over its neighborhoods with different size, which helps to recognize the correlation among node representations of different propagation steps.
This multi-scale feature representation is then fed into the SGA branch to generate the refined attention feature representation for each node.
The SGA branch will gradually remove the noisy level of localities and emphasize those neighborhood regions that are more relevant to the semantic descriptions of the targets.
These duple branches are capable of modeling a wider neighborhood while enhancing correlations, which brings a better feature representation for each node.

\paragraph{\underline{Model Training}}
We combine the loss of two branches as follows:
\begin{equation}
\small
\mathcal{L}=\alpha_t\mathcal{L}_{NA}+(1-\alpha_t)\mathcal{L}_{SGA},
\label{eq_ei}
\end{equation} 
where the time-sensitive parameter $\alpha_t=\cos\left(\frac{\pi t}{2T}\right)$ gives higher weight to the NA branch to find a better reference vector at the beginning and gradually shift the focus to the SGA branch for adaptive adjustment as the training process progresses.

\begin{algorithm}[t]
  \SetAlgoLined
  \KwIn{Graph $G$, propagation number $K$, feature matrix $\mathbf{X}$.}
  \SetAlgoLined
  \caption{\sys-GU}
  \label{alg:sgc}
   $M =\{\mathbf{X}\}$\;
  \For{$1\leq t\leq T$}
  {
    \For{$v\in V$}
    {
        $m^t_v \gets$ \texttt{graph\_aggregator}$(m^{t-1}_{\mathcal{N}_v})$\;
    }
    }
    \For{$v\in V$}
    {
    $h_v \gets $\texttt{update}$(m^T_v)$\; 
  }
\end{algorithm}

\begin{algorithm}[t]
  \SetAlgoLined
  \KwIn{Graph $G$, propagation number $K$, feature matrix $\mathbf{X}$.}
  \SetAlgoLined
  \caption{\sys-GMU}
  \label{alg:gmlp-na}
   $M =\{\mathbf{X}\}$\;
  \For{$1\leq t\leq T$}
  {
    \For{$v\in V$}
    {
        $m^t_v \gets$ \texttt{graph\_aggregator}$(m^{t-1}_{\mathcal{N}_v})$\;
        $M_v = M_v \cup \{m^t_v\}$\;
    }
    }
    \For{$v\in V$}
    {
    $c_v \gets $\texttt{message\_aggregator}$(M_v,\ null/ \mathbf{s})$\;
    $h_v \gets $\texttt{update}$(c_v)$\; 
  }
\end{algorithm}
\subsection{\sys Variants}
It is worth pointing out that the above GMLP algorithm framework derives a family of scalable model variants to tackle flexible performance-efficiency tradeoff, through either adopting different provided graph/message aggregators or dropping certain message aggregators. Both adaptive and non-adaptive message aggregators are included in our general GMLP model. We wish to open up several interesting special cases of our model. 

\paragraph{\underline{\sys-GU}} 
This is the most simplified variant, the sequence of operations is reduce to ``\texttt{graph\_aggregator}$\rightarrow$\texttt{update}'', as shown in Algorithm~\ref{alg:sgc}.
No message aggregators are performed and features are aggregated over a single scale $\mathbf{m^T}$ for applying message. However, the \sys-GU still allows a variety of graph aggregators provided in Section~\ref{sec:ga}. 
SGC~\cite{wu2019simplifying} can be taken as a special case of this version with the normalized-adjacency graph aggregator. This variant has the highest efficiency by dropping message aggregators, but it ignores multiple neighborhood scales and their correlations which lead to higher performance. 

\paragraph{\underline{\sys-GMU}}
In this variant, the model performs operations ``\texttt{graph\_aggregator}$\rightarrow$\texttt{message\_aggregator}$\rightarrow$\texttt{update}'', as shown in Algorithm~\ref{alg:gmlp-na}. 
The message aggregators are used to combine node features over the multi-scale neighborhood to improve performance.
If non-adaptive message aggregators are used, the multi-scale messages are indiscriminately aggregated, without effectively exploring their correlation. SIGN~\cite{sign_icml_grl2020} can be viewed as a special case of using the message aggregator of concatenation. 
Moreover, \sys provides a gating aggregator given by equation~\eqref{eq:gate} for node-adaptive message aggregation. However, the reference vector $\mathbf{s}$ is the same for each node, preventing it from unleashing the full potential to capture the correlations between nodes. 
The architecture of \sys-GMU retains the non-adaptive aggregation branch in Figure~\ref{fig:gmlp}, leading to moderate efficiency in \sys model family.

\subsection{Advantages}

\paragraph{\underline{Efficiency}}
As shown in Section \ref{mp}, passing neural messages requires propagating the outputs obtained from neural transformations, i.e., the forward complexity of the neural network inevitably includes the complexity of both state update and propagation. However, as \sys updates the hidden state after the propagation procedure is done, it could perform propagation only once as a pre-processing stage. Therefore, the forward complexity of the overall model is significantly reduced to that of training an MLP, which is $\mathcal{O}(L_uNd^2)$ as shown in Table \ref{algorithm analysis}. 

\paragraph{\underline{Scalability}}
For large graphs that can not be stored locally, each worker requires to gather neighborhood neural messages from shared (distributed) storage, which leads to high communication cost dominating training cycles. 
For example, let $T$ be the number of training epochs, the communication cost of GCN is $\mathcal{O}(L_pMTd)$.
However, since propagation is taken in advance as pre-computation, 
\sys reduces the total communication cost from $\mathcal{O}(L_pMTd))$ to $\mathcal{O}(L_pMd)$, thus scaling to large graphs.

\begin{table}[]
\caption{Algorithm analysis. We denote $N$, $M$ and $d$ as the number of nodes, edges and features respectively. $L_p$ and $L_u$ are the number of graph convolution and update layers, and $k$ refers to the number of sampled nodes in GraphSAGE.} 
\vspace{-2mm}
    \centering
    \resizebox{.98\columnwidth}{!}{
    \begin{tabular}{l|c|c|c|c}
        \toprule
        \textbf{Algorithm} & \textbf{Message typle} & \textbf{Decoupled} & \textbf{Scalable} & \textbf{Forward pass}\\
        \midrule
        GCN~\cite{kipf2016semi} & neural & $\times$ & $\times$ & $\mathcal{O}(L_pMd+L_uNd^2)$\\
        GraphSAGE~\cite{hamilton2017inductive} & neural & $\times$ & $\times$ &  $\mathcal{O}(k^{L_p}Nd^2)$\\
        APPNP~\cite{DBLP:journals/corr/abs-1810-05997} & decoupled neural & \checkmark & $\times$ & $\mathcal{O}(L_pMd+L_uNd^2)$\\
        AP-GCN~\cite{spinelli2020adaptive} & decoupled neural & \checkmark & $\times$ & $\mathcal{O}(L_pMd+L_uNd^2)$\\
        \hline
        SGC & feature & \checkmark & \checkmark & $\mathcal{O}(L_uNd^2)$\\
        \sys-GU & feature & \checkmark & \checkmark & $\mathcal{O}(L_uNd^2)$\\
        \sys-GMU & feature & \checkmark & \checkmark & $\mathcal{O}(L_uNd^2)$\\
        \sys & feature & \checkmark & \checkmark & $\mathcal{O}(L_uNd^2)$\\
        \bottomrule
    \end{tabular}
    }
    
    \label{algorithm analysis}
\end{table}

\section{GMLP Implementation}
Different from the existing GNNs, the training of \sys is clearly separated into a pre-processing stage and a training stage: First, we pre-compute the message vectors for each node over the graph, and then we combine the messages and train the model parameters with SGD. Both stages can be implemented in a distributed fashion.

\begin{figure}[t]
	\centering
	\includegraphics[width=3.5in]{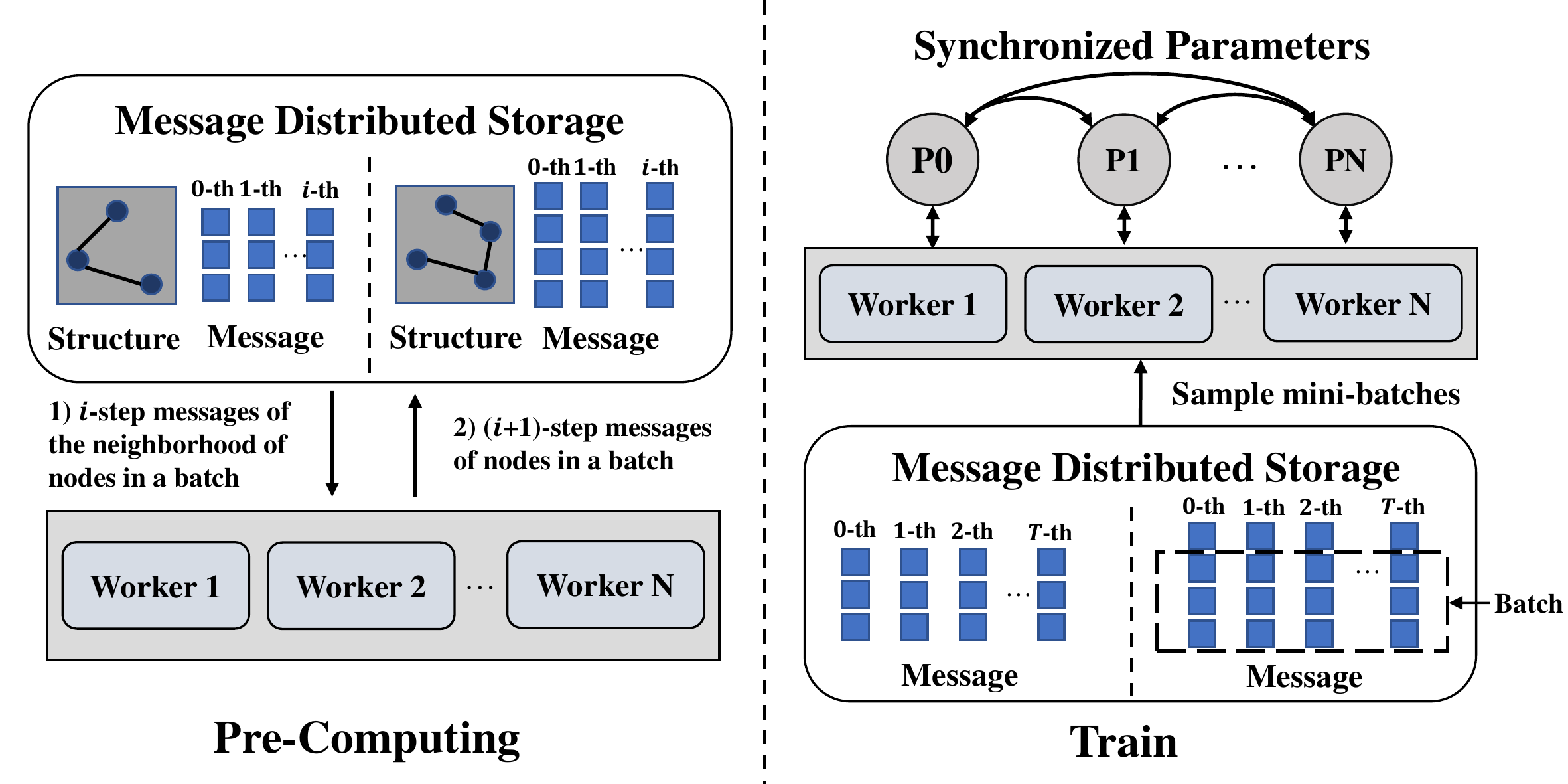}
	\vspace{-4mm}
	\caption{The workflow of GMLP, consisting of message pre-compute stage and message aggregation/update stage}.
	\label{fig:process}
	\vspace{-2mm}
\end{figure}

\paragraph{\emph{Graph message pre-processing.}}
For the first stage, we implement an efficient batch data processing pipeline function over distributed graph storage: The nodes are partitioned into batches, and the computation of each batch is implemented by workers in parallel with matrix multiplication.
As shown in Figure~\ref{fig:process}, for each node in a batch, we firstly pull all the $i$-th step messages of its 1-hop neighbors from the message distributed storage and then compute the $(i+1)$-th step messages of the batch in parallel. Next, We push these aggregated messages back for reuse in the calculation of the $(i+2)$-step messages. 
In our implementation, we treat GPUs as workers for fast pre-processing, and the graph data are partitioned and stored on host memory across machines. 
Since we compute the message vectors for each node in parallel, our implementation could scale to large graphs and significantly reduce the runtime.

\paragraph{\emph{Distributed training.}} 
For the second stage, we implement \sys by PyTorch and optimize the parameters with distributed SGD.
For algorithms that can be expressed by the \sys interfaces, we translate it into the corresponding model architecture containing message aggregator operations such as attention. 
Then, the model parameters are stored on a parameter server and multiple workers (GPU) process the data in parallel. 
We adopt asynchronous training to avoid the communication overhead between many workers. 
Each worker fetches the most up-to-date parameters and computes the gradients
for a mini-batch of data, independent of the other workers. 

\vspace{-2mm}
\begin{table}[h]
\caption{Overview of datasets (T and I represents transductive and inductive, respectively).}
\vspace{-2mm}
\label{Dataset}
\centering
{
\noindent
\renewcommand{\multirowsetup}{\centering}
\resizebox{0.96\linewidth}{!}{
\begin{tabular}{ccccccc}
\toprule
\textbf{Dataset}&\textbf{\#Nodes}& \textbf{\#Features}&\textbf{\#Edges}&\textbf{\#Classes}&\textbf{Task}\\
\midrule
Cora& 2,708 & 1,433 &5,429&7& T \\ 
Citeseer& 3,327 & 3,703&4,732&6& T \\ 
Pubmed& 19,717 & 500 &44,338&3& T\\ 
Amazon Computer& 13,381  & 767& 245,778 & 10 &T\\
Amazon Photo &7,487  & 745& 119,043 & 8 & T\\
Coauthor CS& 18,333  & 6,805 & 81,894 & 15 & T\\
Coauthor Physics& 34,493 & 8,415 & 247,962 & 5 & T\\
ogbn-products&2,449,029 & 100 & 61,859,140 & 47 & T\\
\midrule
Flickr& 89,250 & 500 & 899,756 & 7 & I\\
Reddit& 232,965 & 602 & 11,606,919 & 41 & I\\
\midrule
Tencent & 1,000,000 & 64 & 1,434,382 & 253 & T\\
\bottomrule
\end{tabular}}}
\label{data}
\vspace{-5mm}
\end{table}

\begin{table*}[t]
\caption{Test accuracy (\%) in transductive settings (Dcp. neural means decoupled neural).} 
\vspace{-2mm}
\centering
{
\noindent
\renewcommand{\multirowsetup}{\centering}
\resizebox{0.9\linewidth}{!}{
\begin{tabular}{cccccccccc}
\toprule
\textbf{Type}&\textbf{Models}&\textbf{Cora}& \textbf{Citeseer}&\textbf{Pubmed}&{\textbf{\makecell{Amazon \\Computer}}} & 
{\textbf{\makecell{Amazon \\Photo}}} & 
{\textbf{\makecell{Coauthor \\CS}}}&
{\textbf{\makecell{Coauthor \\Physics}}}&
{\textbf{\makecell{Tencent \\Video}}}\\
\midrule
\multirowcell{4}{Neural}&
GCN& 81.8$\pm$0.5 & 70.8$\pm$0.5 &79.3$\pm$0.7&82.4$\pm$0.4 & 91.2$\pm$0.6 & 90.7$\pm$0.2 & 92.7$\pm$1.1& 45.9$\pm$0.4 \\
&GAT& 83.0$\pm$0.7 & 72.5$\pm$0.7 &79.0$\pm$0.3&80.1$\pm$0.6 & 90.8$\pm$1.0 & 87.4$\pm$0.2 & 90.2$\pm$1.4& 46.8$\pm$0.7 \\
&JK-Net& 81.8$\pm$0.5  & 70.7$\pm$0.7 & 78.8$\pm$0.7 & 82.0$\pm$0.6 & 91.9$\pm$0.7 & 89.5$\pm$0.6 & 92.5$\pm$0.4& 47.2$\pm$0.3 \\
&ResGCN& 82.2$\pm$0.6 & 70.8$\pm$0.7 & 78.3$\pm$0.6& 81.1$\pm$0.7 & 91.3$\pm$0.9 & 87.9$\pm$0.6 & 92.2$\pm$1.5& 46.8$\pm$0.5 \\
\midrule
\multirowcell{2}{Dcp. neural}&
APPNP& 83.3$\pm$0.5 & 71.8$\pm$0.5 & 80.1$\pm$0.2&81.7$\pm$0.3&91.4$\pm$0.3&92.1$\pm$0.4&92.8$\pm$0.9 &46.7$\pm$0.6\\
&AP-GCN& 83.4$\pm$0.3& 71.3$\pm$0.5& 79.7$\pm$0.3&83.7$\pm$0.6& 92.1$\pm$0.3& 91.6$\pm$0.7& 93.1$\pm$0.9&46.9$\pm$0.7\\
\midrule
\multirowcell{2}{Feature}&
SGC & 81.0$\pm$0.2 & 71.3$\pm$0.5 & 78.9$\pm$0.5&82.2$\pm$0.9&91.6$\pm$0.7&90.3$\pm$0.5& 91.7$\pm$1.1 &45.2$\pm$0.3\\
&SIGN& 82.1$\pm$0.3 & 72.4$\pm$0.8 &79.5$\pm$0.5&83.1$\pm$0.8&91.7$\pm$0.7&91.9$\pm$0.3& 92.8$\pm$0.8  &46.3$\pm$0.5\\
\midrule
\multirowcell{3}{Feature}&
GMLP& \textbf{84.1$\pm$0.5} & \textbf{72.7$\pm$0.4} &\textbf{80.3$\pm$0.6}&\textbf{84.7$\pm$0.7}&\textbf{92.6$\pm$0.8}  &\textbf{92.5$\pm$0.5}& \textbf{93.7$\pm$0.9}&\textbf{47.7$\pm$0.3}\\ 
&GMLP-GU(PPR)&81.7$\pm$0.6&71.1$\pm$0.5&79.1$\pm$0.6&82.5$\pm$0.8&91.7$\pm$0.8&90.3$\pm$0.3& 91.4$\pm$0.7&45.7$\pm$0.4\\
&GMLP-GMU(Gating)&83.6$\pm$0.3&72.1$\pm$0.3&79.8$\pm$0.7&83.8$\pm$0.6&91.8$\pm$0.6&91.6$\pm$0.4& 93.1$\pm$0.7&47.4$\pm$0.6\\
\bottomrule
\end{tabular}}}
\vspace{-2mm}
\label{Node1}
\end{table*}

\begin{table}[h]
\caption{ Test accuracy (\%) in inductive settings.}
\vspace{-2mm}
\centering
{
\noindent
\renewcommand{\multirowsetup}{\centering}
\resizebox{0.7\linewidth}{!}{
\begin{tabular}{cccc}
\toprule
\textbf{Models}& \textbf{Flickr}&\textbf{Reddit}\\
\midrule
GraphSAGE & 50.1$\pm$1.3 & 95.4$\pm$0.0 \\
FastGCN & 50.4$\pm$0.1 & 93.7$\pm$0.0 \\
ClusterGCN & 48.1$\pm$0.5 & 95.7$\pm$0.0 \\
GraphSAINT & 51.1$\pm$0.1 & \textbf{96.6$\pm$0.1}\\
\midrule
GMLP& \textbf{52.3$\pm$0.2}  & \textbf{96.6$\pm$0.1}  \\
GMLP-GU(PPR)&50.3$\pm$0.3&95.2$\pm$0.1\\
GMLP-GMU(Gating)&51.6$\pm$0.2&96.1$\pm$0.0\\
\bottomrule
\end{tabular}}}
\label{Inductive}
\end{table}

\section{Experiments}
\subsection{Experimental Settings}
\noindent\textbf{Datasets.}
We conduct the experiments on public partitioned datasets, including three citation networks (Citeseer, Cora, and Pubmed) in~\cite{DBLP:conf/iclr/KipfW17}, two social networks (Flickr and Reddit) in~\cite{DBLP:conf/iclr/ZengZSKP20}, four co-authorship graphs (Amazon and Coauthor) in~\cite{DBLP:conf/iclr/PeiWCLY20}, the co-purchasing network (ogbn-products) in~\cite{DBLP:conf/nips/HuFZDRLCL20} and the tencent video dataset from our industry partner --- Tencent Inc.
Table~\ref{data} provides the overview of the 11 datasets and the detailed description is in Section A.1 of the supplemental material.

\noindent\textbf{Parameters.}
For GMLP and other baselines, we use random search or follow the original papers to get the optimal hyperparameters.
To eliminate random factors, we run each method 20 times and report the mean and variance of the performance. 
More details can be found in Sec. A.3 in the supplementary material.

\noindent\textbf{Environment.} We run our experiments on four machines, each with 14 Intel(R) Xeon(R) CPUs (Gold 5120 @ 2.20GHz) and 4 NVIDIA TITAN RTX GPUs.
All the experiments are implemented in Python 3.6 with Pytorch 1.7.1 on CUDA 10.1.

\noindent\textbf{Baselines.}
In the transductive settings, we compare GMLP with GCN~\cite{DBLP:conf/iclr/KipfW17}, GAT~\cite{DBLP:conf/iclr/VelickovicCCRLB18}, JK-Net~\cite{DBLP:conf/icml/XuLTSKJ18}, Res-GCN~\cite{DBLP:conf/iclr/KipfW17}, APPNP~\cite{DBLP:conf/iclr/KlicperaBG19}, AP-GCN~\cite{spinelli2020adaptive}, SGC~\cite{wu2019simplifying}, SIGN~\cite{DBLP:journals/corr/abs-2004-11198}, which are the state-of-the-art models of different message passing types. Besides, we also compare GMLP with its two variants: GMLP-GU with the personalized pagerank graph aggregators and GMLP-GMU with the gating message aggregators.
In the inductive settings, the compared baselines are GraphSAGE~\cite{hamilton2017inductive}, FastGCN~\cite{DBLP:conf/iclr/ChenMX18}, ClusterGCN~\cite{DBLP:conf/kdd/ChiangLSLBH19} and GraphSAINT~\cite{DBLP:conf/iclr/ZengZSKP20}. The detailed introduction of these baselines are shown in Section A.2 of the supplemental material.

\begin{figure}[tpb]
    \centering
    \includegraphics[width=0.8\linewidth]{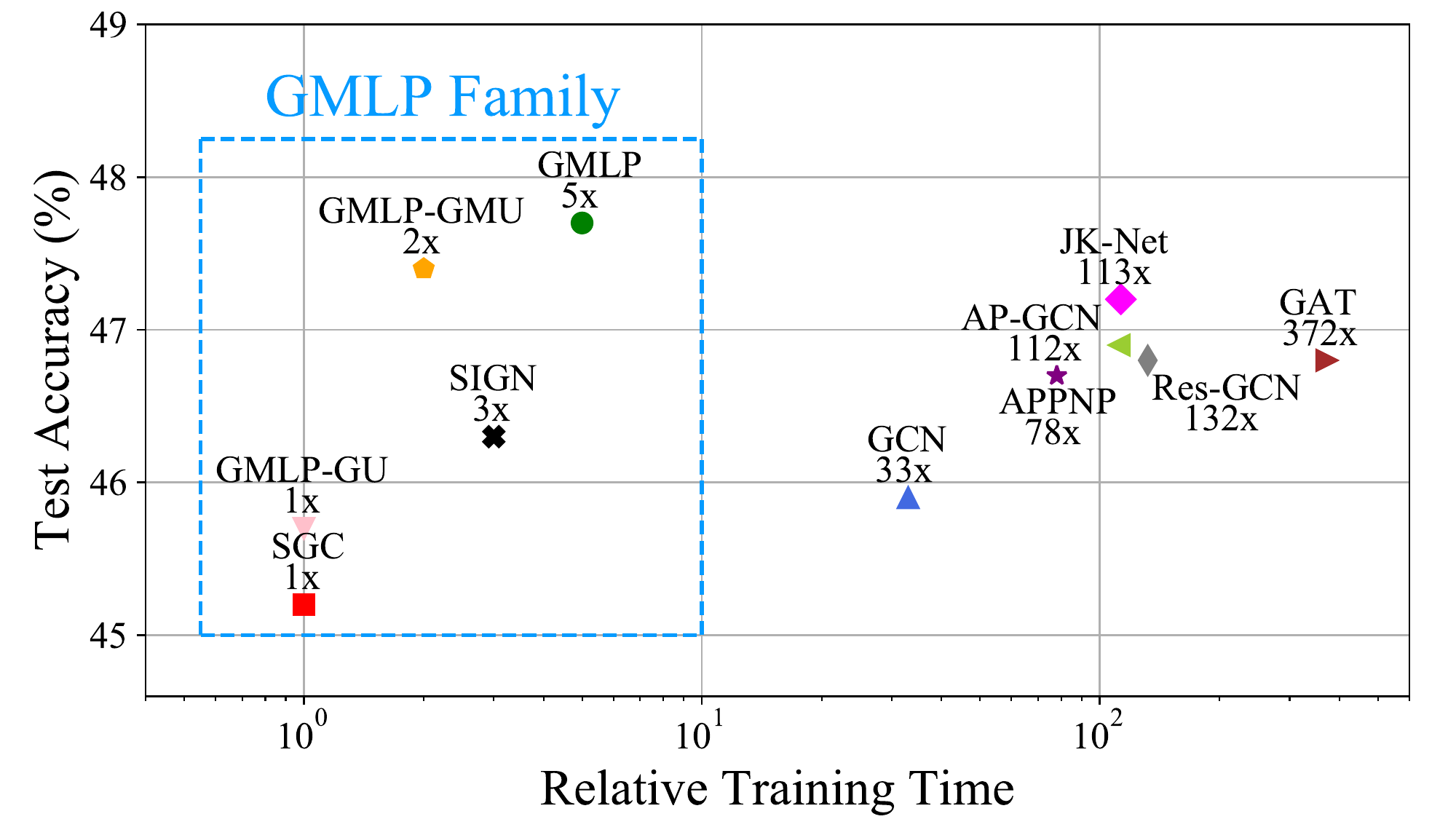}
    \vspace{-2mm}
    \caption{Performance over training time on Tencent Video.}
    \label{acc-efficiency}
\end{figure}

\subsection{Performance-Efficiency Analysis}
To demonstrate the overall performance in both transductive and inductive settings, we compare GMLP with the other state-of-the-art methods. The results are summarized in Table~\ref{Node1} and \ref{Inductive}. 

We observe that \sys obtains quite competitive performance in both transductive and inductive settings. In inductive settings, Table \ref{Inductive} shows that \sys outperforms the best baseline GraphSAINT by a margin of 1.2\% on Flickr while achieves the same performance on Reddit. In transductive settings, \sys outperforms the best baseline of each dataset by a margin of 0.3\% to 1.0\%.
Remarkably, our simplified variant \sys-GMU also achieves the best performance among the baselines that pass neural messages, which shows the superiority of passing node features followed by message aggregation. 
In addition, \sys improves SIGN, the best scalable baselines, by a margin of 0.3\% to 2.0\%. We attribute this improvement to the application of the self-guided adaptive message aggregator.

We also evaluate the efficiency of each method in the real production environment.
Figure \ref{acc-efficiency} illustrates the performance over training time on Tencent Video. In particular, we pre-compute the graph messages of each scalable method, and the training time takes into account the pre-computation time. 
We observe that GCN, along with its variants which pass neural messages requires a rather larger training time than the variants of \sys that pass node features. 
Among considered baselines, \sys achieves the best performance with 5$\times$ training time compared with \sys-GU and SGC. 
It's worth pointing out that \sys-GU and \sys-GMU varients, outperform SGC and SIGN respectively while requiring less training time.

\begin{figure*}[htbp]
\centering
\subfigure[Stand-alone on Reddit]{
    \begin{minipage}[t]{0.24\linewidth}
    \centering
    \includegraphics[width=0.85\linewidth]{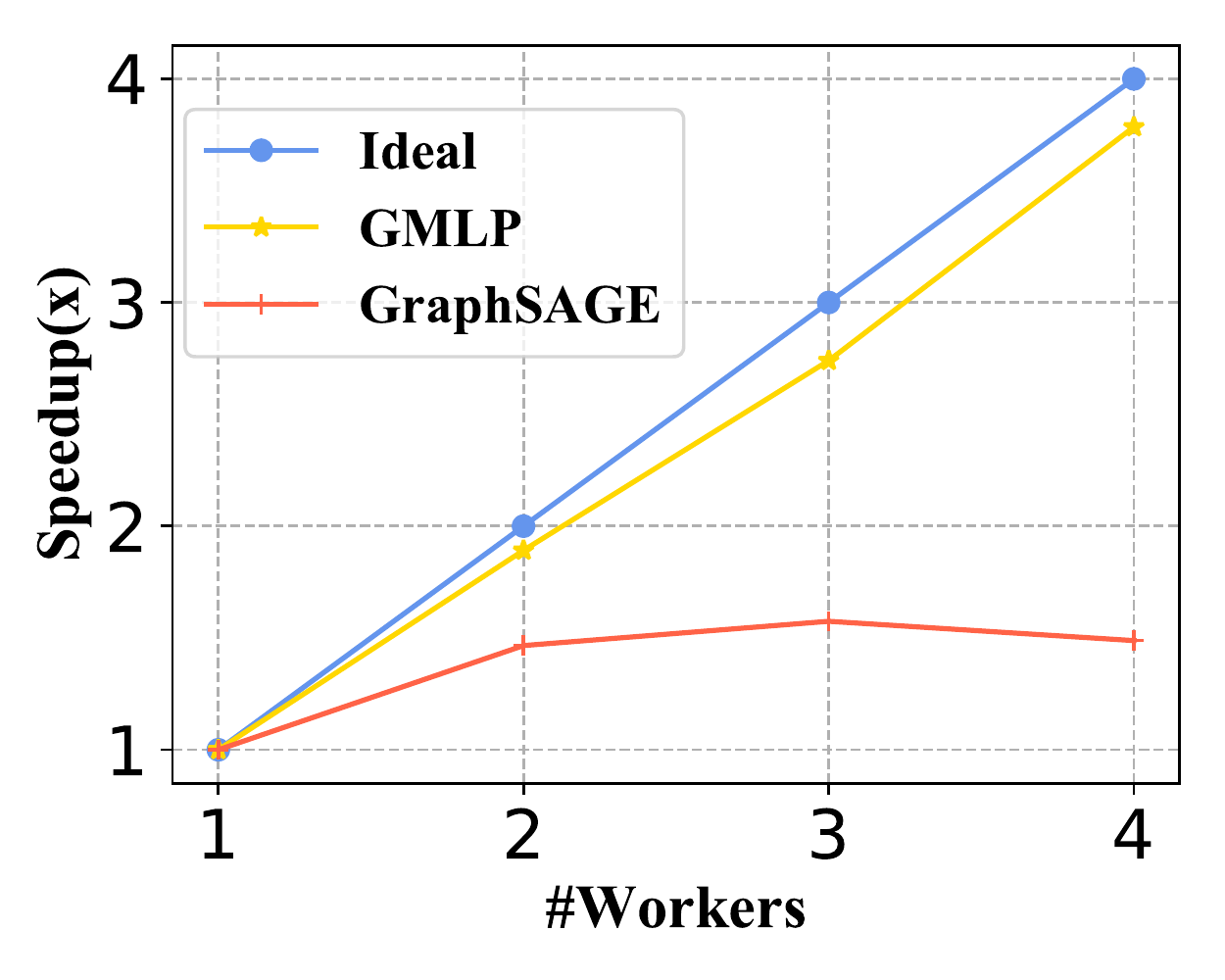}
    \end{minipage}%
    }
\subfigure[Distributed on Reddit]{
    \begin{minipage}[t]{0.24\linewidth}
    \centering
    \includegraphics[width=0.85\linewidth]{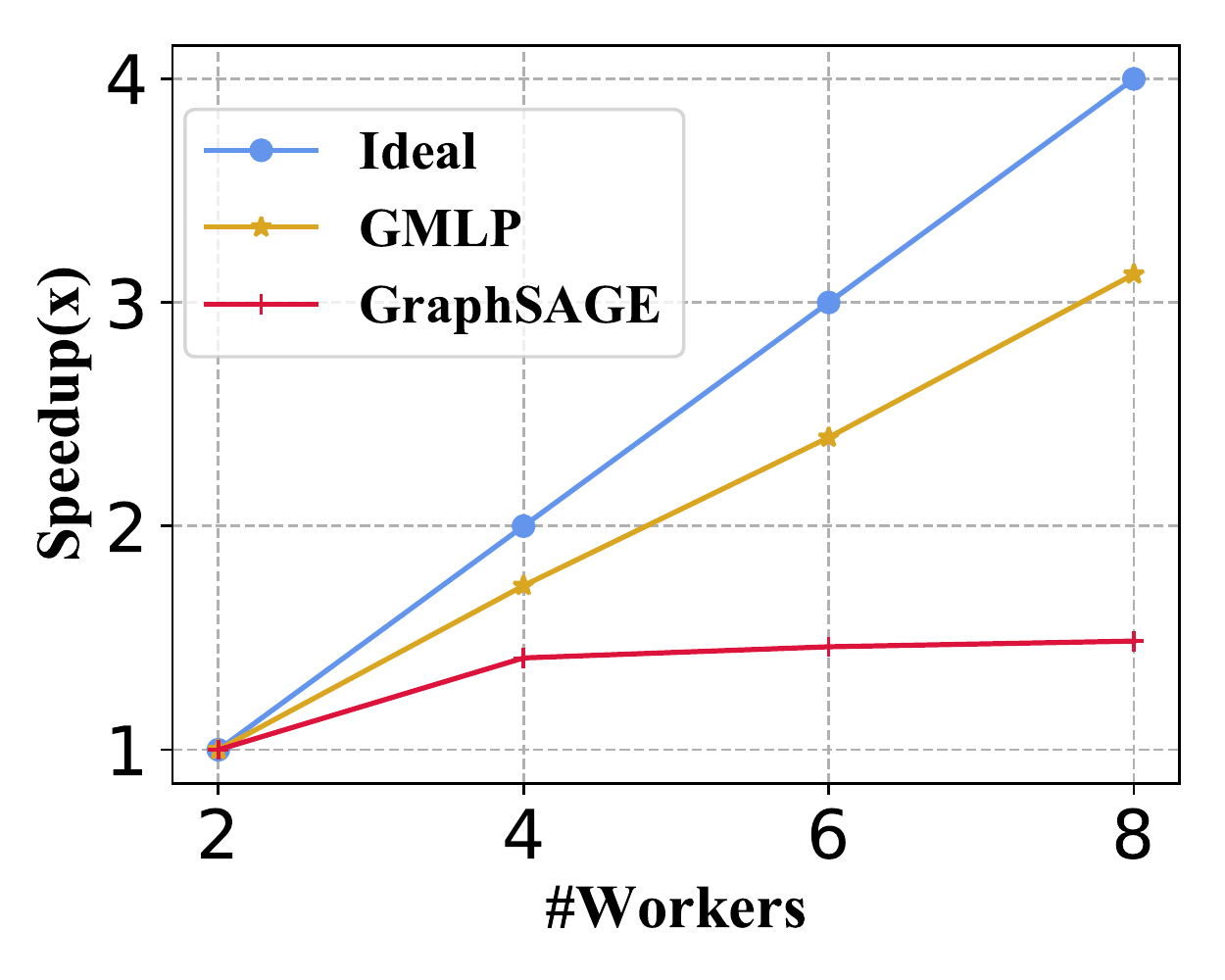}
    \end{minipage}%
    }%
\subfigure[Stand-alone on ogbn-product]{
    \begin{minipage}[t]{0.24\linewidth}
    \centering
    \includegraphics[width=0.85\linewidth]{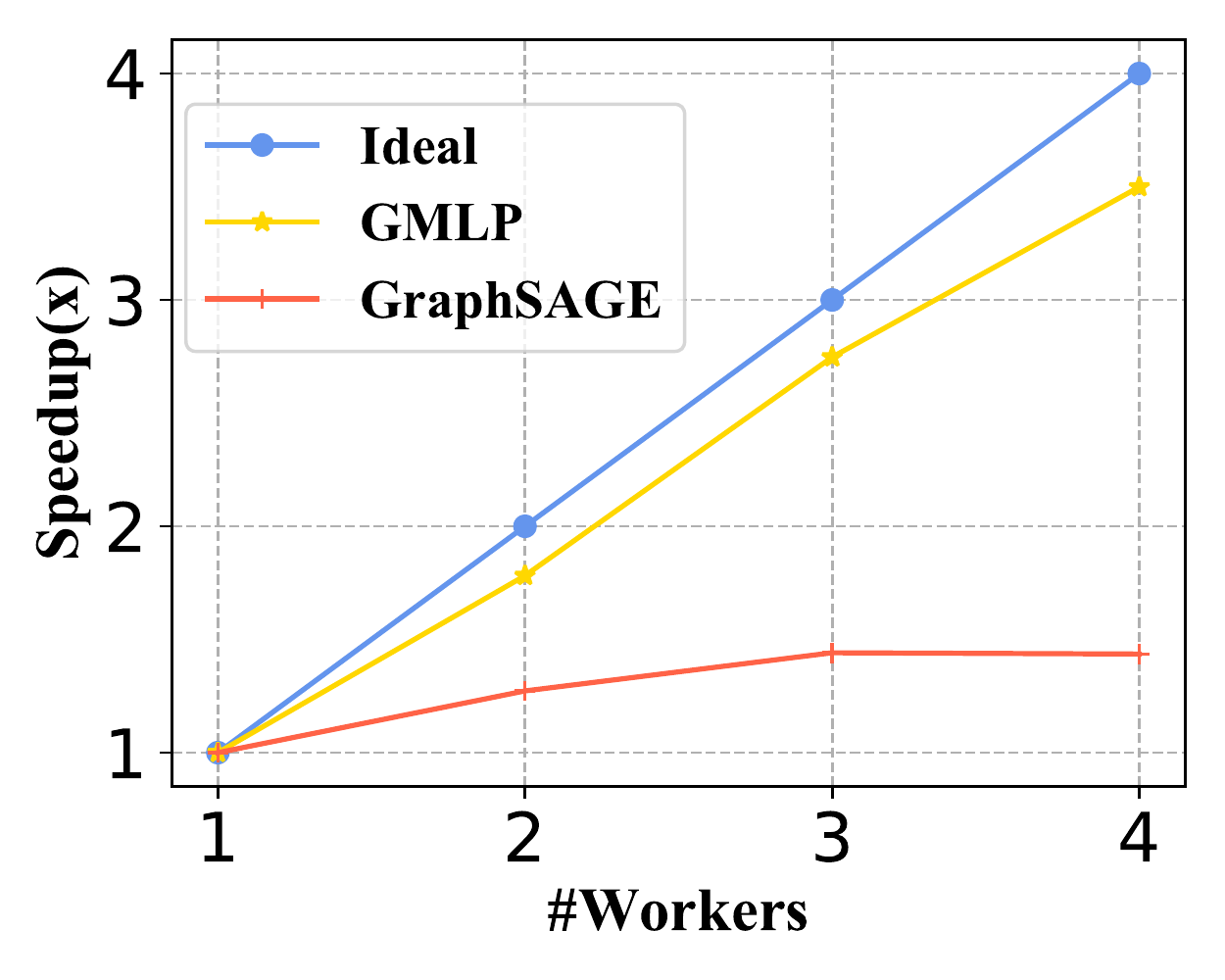}
    \end{minipage}
    }%
\subfigure[Distributed on ogbn-product]{
    \begin{minipage}[t]{0.24\linewidth}
    \centering
    \includegraphics[width=0.85\linewidth]{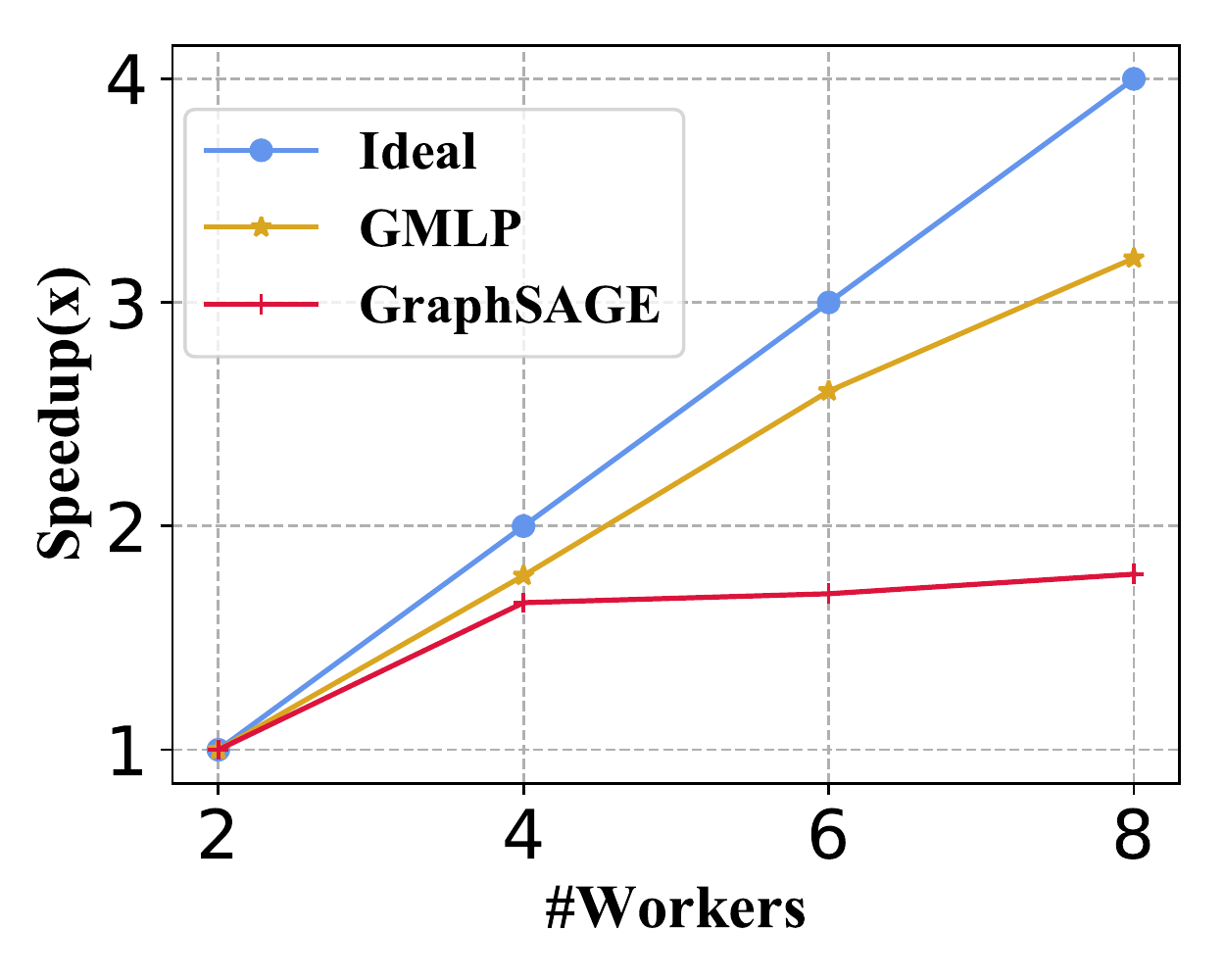}
    \end{minipage}
    }%
\vspace{-4mm}
\centering
\caption{Scalability comparison on Reddit and ogbn-product datasets. The stand-alone scenario means the graph has only one partition stored on a multi-GPU server, whereas the distributed scenario means the graph is partitioned and stored on multi-servers. In the distributed scenario, we run two workers per machine.}
\label{scalability}
\vspace{-4mm}
\end{figure*}

\subsection{Training Scalability}
To examine the training scalability of \sys, we compare it with GraphSAGE, a widely used method in industry on two large-scale datasets, and the results are shown in Figure~\ref{scalability}.
We run the two methods in both stand-alone and distributed scenarios, and then measure their corresponding speedups. 
The batch size is set to 8192 for Reddit and 16384 for ogbn-product, and the speedup is calculated by runtime per epoch relative to that of one worker in the stand-alone scenario and two workers in the distributed scenario.
Without considering extra costs, the speedup will increase linearly in an ideal condition. For GraphSAGE, since it requires aggregating the neighborhood nodes during training, it meets the I/O bottleneck when transmitting a large number of required neural messages. Thus, the speedup of GraphSAGE increases slowly as the number of workers grows. The speedup of GraphSAGE is less than 2$\times$ even with 4 workers in the stand-alone scenario and 8 workers in the distributed scenario.
It's worth recalling that the only extra communication cost of GMLP is to synchronize parameters with different workers, which is essential to all distributed training methods. As a result, \sys performs more scalable than GraphSAGE and behaves close to the ideal circumstance in both two scenarios.

\begin{figure}[tpb]
    \centering
    \includegraphics[width=0.8\linewidth]{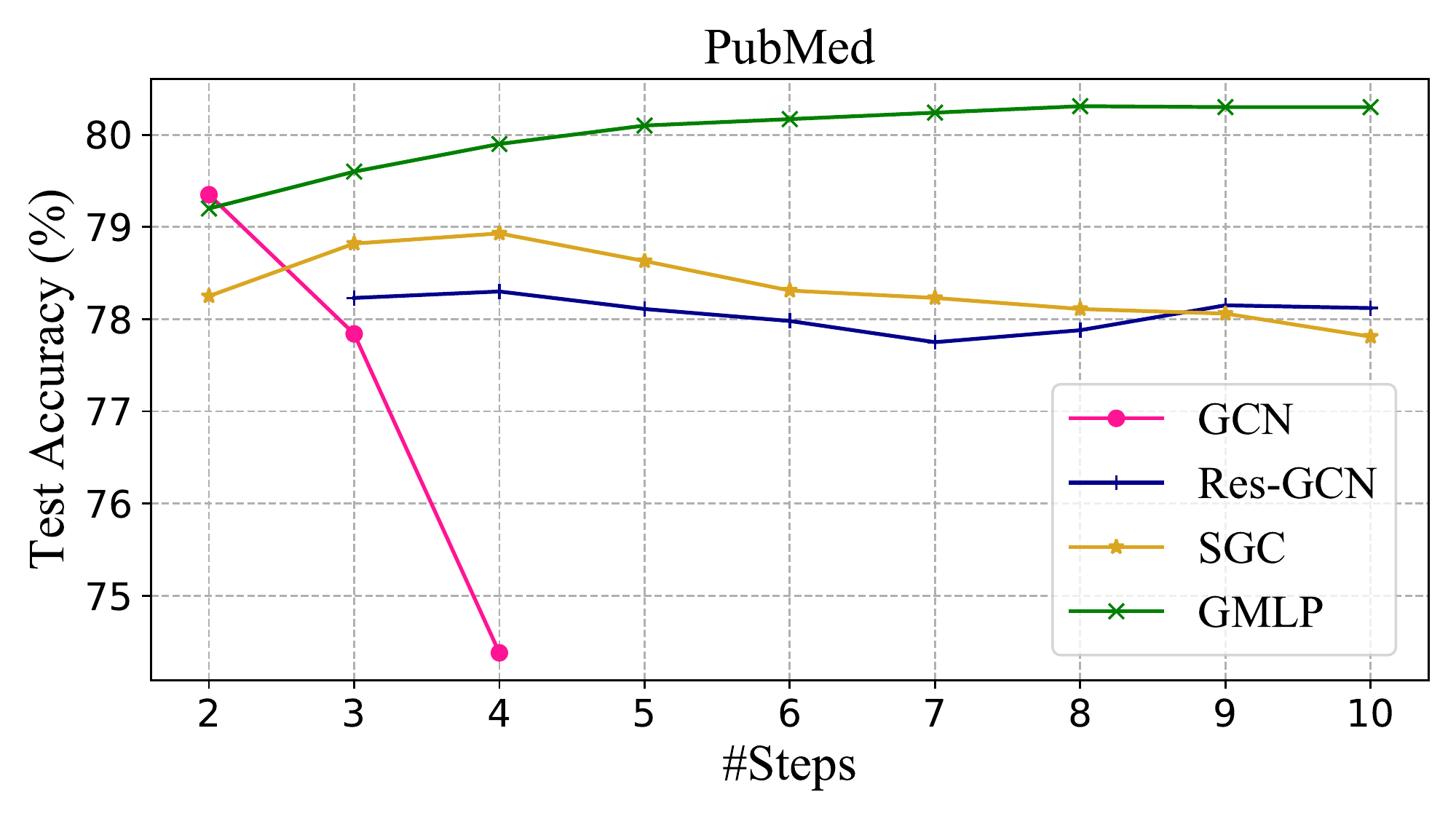}
    \vspace{-4mm}
    \caption{Test accuracy of different models along with the increased aggregation steps.}
    \label{depth}
\end{figure}

\vspace{-4mm}
\subsection{Model Scalability}
We examine the model scalability by observing how the model performance changes along with the message passing step $T$.
As shown in Figure~\ref{depth}, the vanilla GCN gets the best results with two aggregation steps, but its performance drops rapidly along with the increased steps due to the over-smoothing issue.
Both ResGCN and SGC show better performance than GCN with larger aggregation steps. 
SGC alleviates this problem by removing the non-linear transformation and ResGCN carries information from the previous step by introducing the residual connections.
However, their performance still degrades as they are unable to balance the needs of preserving locality (i.e. staying close to the root node to avoid over-smoothing) and leveraging the information from a large neighborhood. 
In contrast, GMLP achieves consistent performance improvement across steps, which indicates that GMLP is able to adaptively and effectively combine multi-scale neighborhood messages for each node.

\begin{figure}[tpb]
    \centering
    \includegraphics[width=0.8\columnwidth, height=1.1in]{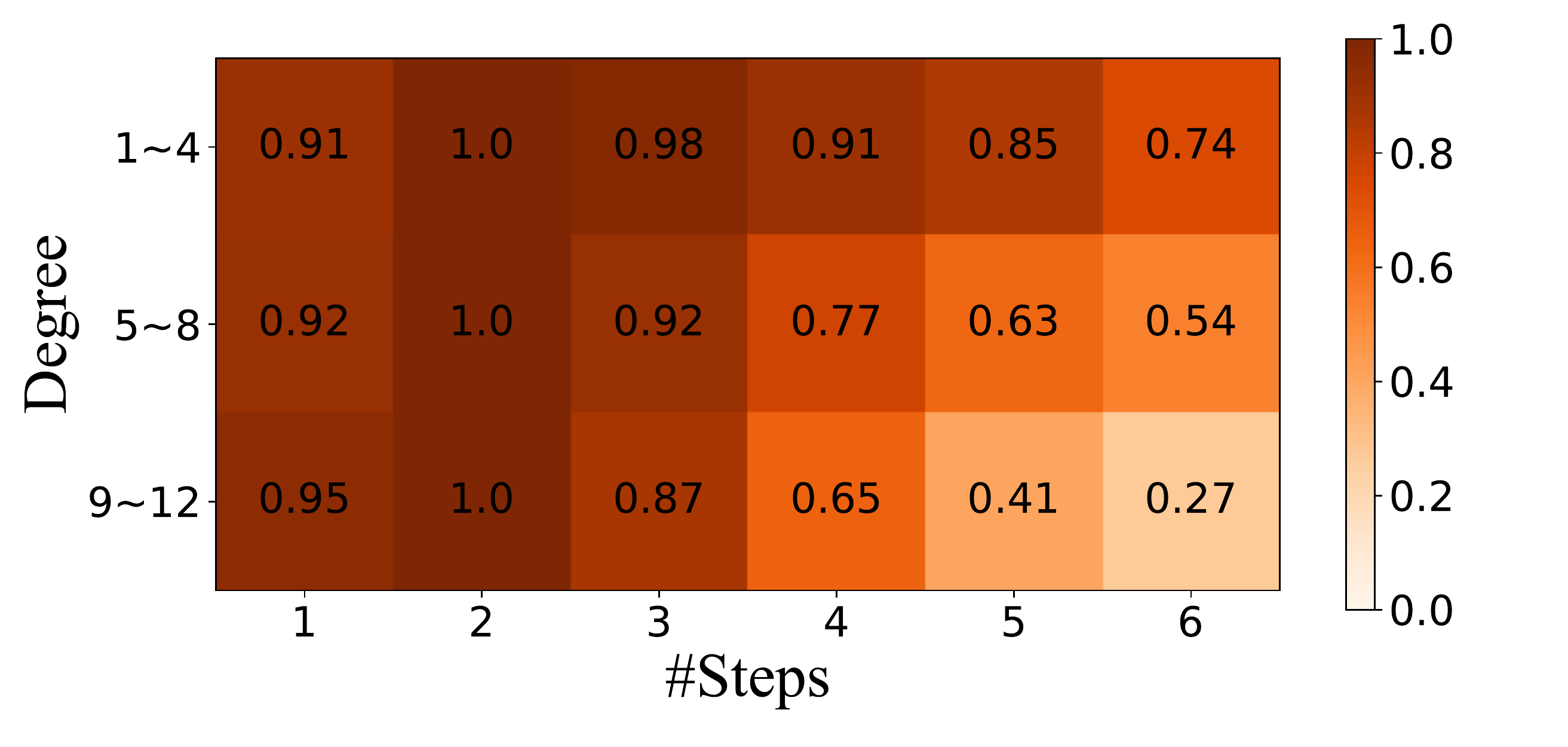}
    \vspace{-4mm}
    \caption{The average attention weights of graph messages of different steps on 60 randomly selected nodes from Cora. }
    \label{interpretability}
\end{figure}

To demonstrate this, Figure \ref{interpretability} shows the average attention weights of graph messages according to the number of steps and degrees of input nodes, where the maximum step is 6. In this experiment, we randomly select 20 nodes for each degree range (1-4, 5-8, 9-12) and plot the relative weight based on the maximum value. 
We get two observations from the heat map: 1) The 1-step and 2-step graph messages are always of great importance, which shows that \sys captures the local information as those widely 2-layer methods do; 2) The weights of graph messages with larger steps drop faster as the degree grows, which indicates that the attention-based aggregator could prevent high-degree nodes from including excessive irrelevant nodes which lead to over-smoothing. From the two observations, we conclude that \sys is able to identify the different message passing demands of nodes and explicitly weight each graph message.




\section{Conclusion}
We present \sys, a new GNN framework that achieves the best of both worlds of scalability and performance via a novel feature message passing abstraction. 
In comparison to previous approaches, \sys decouples message passing and neural update and solves the limited scalability and flexibility problems inherent in previous neural message passing models.
Under this abstraction, \sys provides a variety of graph and message aggregators as cornerstones for developing scalable GNNs. By exploring different aggregators under the framework, we derive novel \sys model variants that allow for efficient feature aggregation over adaptive node neighborhoods.
Experiments on 11 real-world benchmark datasets demonstrate that \sys consistently outperforms the other state-of-the-art methods on performance, efficiency, and scalability.
For future work, we are building GMLP and integrating it as a key part of Angel-Graph~\footnote{\url{https://github.com/Angel-ML/angel}}.

\normalem
\bibliographystyle{ACM-Reference-Format}
\bibliography{reference}


\begin{thebibliography}{39}


\ifx \showCODEN    \undefined \def \showCODEN     #1{\unskip}     \fi
\ifx \showDOI      \undefined \def \showDOI       #1{#1}\fi
\ifx \showISBNx    \undefined \def \showISBNx     #1{\unskip}     \fi
\ifx \showISBNxiii \undefined \def \showISBNxiii  #1{\unskip}     \fi
\ifx \showISSN     \undefined \def \showISSN      #1{\unskip}     \fi
\ifx \showLCCN     \undefined \def \showLCCN      #1{\unskip}     \fi
\ifx \shownote     \undefined \def \shownote      #1{#1}          \fi
\ifx \showarticletitle \undefined \def \showarticletitle #1{#1}   \fi
\ifx \showURL      \undefined \def \showURL       {\relax}        \fi
\providecommand\bibfield[2]{#2}
\providecommand\bibinfo[2]{#2}
\providecommand\natexlab[1]{#1}
\providecommand\showeprint[2][]{arXiv:#2}

\bibitem[\protect\citeauthoryear{Berg, Kipf, and Welling}{Berg
  et~al\mbox{.}}{2017}]%
        {berg2017graph}
\bibfield{author}{\bibinfo{person}{Rianne van~den Berg},
  \bibinfo{person}{Thomas~N Kipf}, {and} \bibinfo{person}{Max Welling}.}
  \bibinfo{year}{2017}\natexlab{}.
\newblock \showarticletitle{Graph convolutional matrix completion}.
\newblock \bibinfo{journal}{\emph{arXiv preprint arXiv:1706.02263}}
  (\bibinfo{year}{2017}).
\newblock


\bibitem[\protect\citeauthoryear{Bronstein, Bruna, LeCun, Szlam, and
  Vandergheynst}{Bronstein et~al\mbox{.}}{2017}]%
        {bronstein2017geometric}
\bibfield{author}{\bibinfo{person}{Michael~M Bronstein}, \bibinfo{person}{Joan
  Bruna}, \bibinfo{person}{Yann LeCun}, \bibinfo{person}{Arthur Szlam}, {and}
  \bibinfo{person}{Pierre Vandergheynst}.} \bibinfo{year}{2017}\natexlab{}.
\newblock \showarticletitle{Geometric deep learning: going beyond euclidean
  data}.
\newblock \bibinfo{journal}{\emph{IEEE Signal Processing Magazine}}
  \bibinfo{volume}{34}, \bibinfo{number}{4} (\bibinfo{year}{2017}),
  \bibinfo{pages}{18--42}.
\newblock


\bibitem[\protect\citeauthoryear{Cai and Ji}{Cai and Ji}{2020}]%
        {cai2020multi}
\bibfield{author}{\bibinfo{person}{Lei Cai} {and} \bibinfo{person}{Shuiwang
  Ji}.} \bibinfo{year}{2020}\natexlab{}.
\newblock \showarticletitle{A multi-scale approach for graph link prediction}.
  In \bibinfo{booktitle}{\emph{Proceedings of the AAAI Conference on Artificial
  Intelligence}}, Vol.~\bibinfo{volume}{34}. \bibinfo{pages}{3308--3315}.
\newblock


\bibitem[\protect\citeauthoryear{Chen, Lin, Li, Li, Zhou, and Sun}{Chen
  et~al\mbox{.}}{2020}]%
        {chen2020measuring}
\bibfield{author}{\bibinfo{person}{Deli Chen}, \bibinfo{person}{Yankai Lin},
  \bibinfo{person}{Wei Li}, \bibinfo{person}{Peng Li}, \bibinfo{person}{Jie
  Zhou}, {and} \bibinfo{person}{Xu Sun}.} \bibinfo{year}{2020}\natexlab{}.
\newblock \showarticletitle{Measuring and relieving the over-smoothing problem
  for graph neural networks from the topological view}. In
  \bibinfo{booktitle}{\emph{Proceedings of the AAAI Conference on Artificial
  Intelligence}}, Vol.~\bibinfo{volume}{34}. \bibinfo{pages}{3438--3445}.
\newblock


\bibitem[\protect\citeauthoryear{Chen, Ma, and Xiao}{Chen
  et~al\mbox{.}}{2018}]%
        {DBLP:conf/iclr/ChenMX18}
\bibfield{author}{\bibinfo{person}{Jie Chen}, \bibinfo{person}{Tengfei Ma},
  {and} \bibinfo{person}{Cao Xiao}.} \bibinfo{year}{2018}\natexlab{}.
\newblock \showarticletitle{FastGCN: Fast Learning with Graph Convolutional
  Networks via Importance Sampling}. In \bibinfo{booktitle}{\emph{6th
  International Conference on Learning Representations, {ICLR} 2018, Vancouver,
  BC, Canada, April 30 - May 3, 2018, Conference Track Proceedings}}.
  \bibinfo{publisher}{OpenReview.net}.
\newblock


\bibitem[\protect\citeauthoryear{Chiang, Liu, Si, Li, Bengio, and Hsieh}{Chiang
  et~al\mbox{.}}{2019}]%
        {DBLP:conf/kdd/ChiangLSLBH19}
\bibfield{author}{\bibinfo{person}{Wei{-}Lin Chiang}, \bibinfo{person}{Xuanqing
  Liu}, \bibinfo{person}{Si Si}, \bibinfo{person}{Yang Li},
  \bibinfo{person}{Samy Bengio}, {and} \bibinfo{person}{Cho{-}Jui Hsieh}.}
  \bibinfo{year}{2019}\natexlab{}.
\newblock \showarticletitle{Cluster-GCN: An Efficient Algorithm for Training
  Deep and Large Graph Convolutional Networks}. In
  \bibinfo{booktitle}{\emph{Proceedings of the 25th {ACM} {SIGKDD}
  International Conference on Knowledge Discovery {\&} Data Mining, {KDD} 2019,
  Anchorage, AK, USA, August 4-8, 2019}},
  \bibfield{editor}{\bibinfo{person}{Ankur Teredesai}, \bibinfo{person}{Vipin
  Kumar}, \bibinfo{person}{Ying Li}, \bibinfo{person}{R{\'{o}}mer Rosales},
  \bibinfo{person}{Evimaria Terzi}, {and} \bibinfo{person}{George Karypis}}
  (Eds.). \bibinfo{publisher}{{ACM}}, \bibinfo{pages}{257--266}.
\newblock


\bibitem[\protect\citeauthoryear{Frasca, Rossi, Eynard, Chamberlain, Bronstein,
  and Monti}{Frasca et~al\mbox{.}}{2020}]%
        {sign_icml_grl2020}
\bibfield{author}{\bibinfo{person}{Fabrizio Frasca}, \bibinfo{person}{Emanuele
  Rossi}, \bibinfo{person}{Davide Eynard}, \bibinfo{person}{Benjamin
  Chamberlain}, \bibinfo{person}{Michael Bronstein}, {and}
  \bibinfo{person}{Federico Monti}.} \bibinfo{year}{2020}\natexlab{}.
\newblock \showarticletitle{SIGN: Scalable Inception Graph Neural Networks}. In
  \bibinfo{booktitle}{\emph{ICML 2020 Workshop on Graph Representation Learning
  and Beyond}}.
\newblock


\bibitem[\protect\citeauthoryear{Gao, Wang, and Ji}{Gao et~al\mbox{.}}{2018}]%
        {gao2018large}
\bibfield{author}{\bibinfo{person}{Hongyang Gao}, \bibinfo{person}{Zhengyang
  Wang}, {and} \bibinfo{person}{Shuiwang Ji}.} \bibinfo{year}{2018}\natexlab{}.
\newblock \showarticletitle{Large-scale learnable graph convolutional
  networks}. In \bibinfo{booktitle}{\emph{Proceedings of the 24th ACM SIGKDD
  International Conference on Knowledge Discovery \& Data Mining}}.
  \bibinfo{pages}{1416--1424}.
\newblock


\bibitem[\protect\citeauthoryear{Hamilton, Ying, and Leskovec}{Hamilton
  et~al\mbox{.}}{2017}]%
        {hamilton2017inductive}
\bibfield{author}{\bibinfo{person}{Will Hamilton}, \bibinfo{person}{Zhitao
  Ying}, {and} \bibinfo{person}{Jure Leskovec}.}
  \bibinfo{year}{2017}\natexlab{}.
\newblock \showarticletitle{Inductive representation learning on large graphs}.
  In \bibinfo{booktitle}{\emph{NIPS}}. \bibinfo{pages}{1024--1034}.
\newblock


\bibitem[\protect\citeauthoryear{Hu, Fey, Zitnik, Dong, Ren, Liu, Catasta, and
  Leskovec}{Hu et~al\mbox{.}}{2020}]%
        {DBLP:conf/nips/HuFZDRLCL20}
\bibfield{author}{\bibinfo{person}{Weihua Hu}, \bibinfo{person}{Matthias Fey},
  \bibinfo{person}{Marinka Zitnik}, \bibinfo{person}{Yuxiao Dong},
  \bibinfo{person}{Hongyu Ren}, \bibinfo{person}{Bowen Liu},
  \bibinfo{person}{Michele Catasta}, {and} \bibinfo{person}{Jure Leskovec}.}
  \bibinfo{year}{2020}\natexlab{}.
\newblock \showarticletitle{Open Graph Benchmark: Datasets for Machine Learning
  on Graphs}. In \bibinfo{booktitle}{\emph{Advances in Neural Information
  Processing Systems 33: Annual Conference on Neural Information Processing
  Systems 2020, NeurIPS 2020, December 6-12, 2020, virtual}}.
\newblock


\bibitem[\protect\citeauthoryear{Kipf and Welling}{Kipf and Welling}{2016}]%
        {kipf2016semi}
\bibfield{author}{\bibinfo{person}{Thomas~N Kipf} {and} \bibinfo{person}{Max
  Welling}.} \bibinfo{year}{2016}\natexlab{}.
\newblock \showarticletitle{Semi-supervised classification with graph
  convolutional networks}.
\newblock \bibinfo{journal}{\emph{arXiv preprint arXiv:1609.02907}}
  (\bibinfo{year}{2016}).
\newblock


\bibitem[\protect\citeauthoryear{Kipf and Welling}{Kipf and Welling}{2017}]%
        {DBLP:conf/iclr/KipfW17}
\bibfield{author}{\bibinfo{person}{Thomas~N. Kipf} {and} \bibinfo{person}{Max
  Welling}.} \bibinfo{year}{2017}\natexlab{}.
\newblock \showarticletitle{Semi-Supervised Classification with Graph
  Convolutional Networks}. In \bibinfo{booktitle}{\emph{ICLR}}.
\newblock


\bibitem[\protect\citeauthoryear{Klicpera, Bojchevski, and
  G{\"{u}}nnemann}{Klicpera et~al\mbox{.}}{2018a}]%
        {DBLP:journals/corr/abs-1810-05997}
\bibfield{author}{\bibinfo{person}{Johannes Klicpera},
  \bibinfo{person}{Aleksandar Bojchevski}, {and} \bibinfo{person}{Stephan
  G{\"{u}}nnemann}.} \bibinfo{year}{2018}\natexlab{a}.
\newblock \showarticletitle{Personalized Embedding Propagation: Combining
  Neural Networks on Graphs with Personalized PageRank}.
\newblock \bibinfo{journal}{\emph{CoRR}}  \bibinfo{volume}{abs/1810.05997}
  (\bibinfo{year}{2018}).
\newblock


\bibitem[\protect\citeauthoryear{Klicpera, Bojchevski, and
  G{\"u}nnemann}{Klicpera et~al\mbox{.}}{2018b}]%
        {klicpera2018predict}
\bibfield{author}{\bibinfo{person}{Johannes Klicpera},
  \bibinfo{person}{Aleksandar Bojchevski}, {and} \bibinfo{person}{Stephan
  G{\"u}nnemann}.} \bibinfo{year}{2018}\natexlab{b}.
\newblock \showarticletitle{Predict then propagate: Graph neural networks meet
  personalized pagerank}.
\newblock \bibinfo{journal}{\emph{arXiv preprint arXiv:1810.05997}}
  (\bibinfo{year}{2018}).
\newblock


\bibitem[\protect\citeauthoryear{Klicpera, Bojchevski, and
  G{\"{u}}nnemann}{Klicpera et~al\mbox{.}}{2019}]%
        {DBLP:conf/iclr/KlicperaBG19}
\bibfield{author}{\bibinfo{person}{Johannes Klicpera},
  \bibinfo{person}{Aleksandar Bojchevski}, {and} \bibinfo{person}{Stephan
  G{\"{u}}nnemann}.} \bibinfo{year}{2019}\natexlab{}.
\newblock \showarticletitle{Predict then Propagate: Graph Neural Networks meet
  Personalized PageRank}. In \bibinfo{booktitle}{\emph{7th International
  Conference on Learning Representations, {ICLR} 2019, New Orleans, LA, USA,
  May 6-9, 2019}}. \bibinfo{publisher}{OpenReview.net}.
\newblock


\bibitem[\protect\citeauthoryear{Lee, Fang, Yeh, and Wang}{Lee
  et~al\mbox{.}}{2018}]%
        {lee2018multi}
\bibfield{author}{\bibinfo{person}{Chung-Wei Lee}, \bibinfo{person}{Wei Fang},
  \bibinfo{person}{Chih-Kuan Yeh}, {and} \bibinfo{person}{Yu-Chiang~Frank
  Wang}.} \bibinfo{year}{2018}\natexlab{}.
\newblock \showarticletitle{Multi-label zero-shot learning with structured
  knowledge graphs}. In \bibinfo{booktitle}{\emph{Proceedings of the IEEE
  conference on computer vision and pattern recognition}}.
  \bibinfo{pages}{1576--1585}.
\newblock


\bibitem[\protect\citeauthoryear{Li, Han, and Wu}{Li et~al\mbox{.}}{2018}]%
        {li2018deeper}
\bibfield{author}{\bibinfo{person}{Qimai Li}, \bibinfo{person}{Zhichao Han},
  {and} \bibinfo{person}{Xiao-Ming Wu}.} \bibinfo{year}{2018}\natexlab{}.
\newblock \showarticletitle{Deeper insights into graph convolutional networks
  for semi-supervised learning}. In \bibinfo{booktitle}{\emph{Proceedings of
  the AAAI Conference on Artificial Intelligence}}, Vol.~\bibinfo{volume}{32}.
\newblock


\bibitem[\protect\citeauthoryear{Lin, Li, Miao, Liu, and Xu}{Lin
  et~al\mbox{.}}{2020}]%
        {lin2020pagraph}
\bibfield{author}{\bibinfo{person}{Zhiqi Lin}, \bibinfo{person}{Cheng Li},
  \bibinfo{person}{Youshan Miao}, \bibinfo{person}{Yunxin Liu}, {and}
  \bibinfo{person}{Yinlong Xu}.} \bibinfo{year}{2020}\natexlab{}.
\newblock \showarticletitle{PaGraph: Scaling GNN training on large graphs via
  computation-aware caching}. In \bibinfo{booktitle}{\emph{Proceedings of the
  11th ACM Symposium on Cloud Computing}}. \bibinfo{pages}{401--415}.
\newblock


\bibitem[\protect\citeauthoryear{Ma, Yang, Miao, Xue, Wu, Zhou, and Dai}{Ma
  et~al\mbox{.}}{2019}]%
        {ma2019neugraph}
\bibfield{author}{\bibinfo{person}{Lingxiao Ma}, \bibinfo{person}{Zhi Yang},
  \bibinfo{person}{Youshan Miao}, \bibinfo{person}{Jilong Xue},
  \bibinfo{person}{Ming Wu}, \bibinfo{person}{Lidong Zhou}, {and}
  \bibinfo{person}{Yafei Dai}.} \bibinfo{year}{2019}\natexlab{}.
\newblock \showarticletitle{Neugraph: parallel deep neural network computation
  on large graphs}. In \bibinfo{booktitle}{\emph{2019 $\{$USENIX$\}$ Annual
  Technical Conference ($\{$USENIX$\}$$\{$ATC$\}$ 19)}}.
  \bibinfo{pages}{443--458}.
\newblock


\bibitem[\protect\citeauthoryear{Marino, Salakhutdinov, and Gupta}{Marino
  et~al\mbox{.}}{2016}]%
        {marino2016more}
\bibfield{author}{\bibinfo{person}{Kenneth Marino}, \bibinfo{person}{Ruslan
  Salakhutdinov}, {and} \bibinfo{person}{Abhinav Gupta}.}
  \bibinfo{year}{2016}\natexlab{}.
\newblock \showarticletitle{The more you know: Using knowledge graphs for image
  classification}.
\newblock \bibinfo{journal}{\emph{arXiv preprint arXiv:1612.04844}}
  (\bibinfo{year}{2016}).
\newblock


\bibitem[\protect\citeauthoryear{Monti, Bronstein, and Bresson}{Monti
  et~al\mbox{.}}{2017}]%
        {monti2017geometric}
\bibfield{author}{\bibinfo{person}{Federico Monti}, \bibinfo{person}{Michael~M
  Bronstein}, {and} \bibinfo{person}{Xavier Bresson}.}
  \bibinfo{year}{2017}\natexlab{}.
\newblock \showarticletitle{Geometric matrix completion with recurrent
  multi-graph neural networks}.
\newblock \bibinfo{journal}{\emph{arXiv preprint arXiv:1704.06803}}
  (\bibinfo{year}{2017}).
\newblock


\bibitem[\protect\citeauthoryear{Monti, Otness, and Bronstein}{Monti
  et~al\mbox{.}}{2018}]%
        {DBLP:journals/corr/abs-1802-01572}
\bibfield{author}{\bibinfo{person}{Federico Monti}, \bibinfo{person}{Karl
  Otness}, {and} \bibinfo{person}{Michael~M. Bronstein}.}
  \bibinfo{year}{2018}\natexlab{}.
\newblock \showarticletitle{MotifNet: a motif-based Graph Convolutional Network
  for directed graphs}.
\newblock \bibinfo{journal}{\emph{CoRR}}  \bibinfo{volume}{abs/1802.01572}
  (\bibinfo{year}{2018}).
\newblock


\bibitem[\protect\citeauthoryear{Pei, Wei, Chang, Lei, and Yang}{Pei
  et~al\mbox{.}}{2020}]%
        {DBLP:conf/iclr/PeiWCLY20}
\bibfield{author}{\bibinfo{person}{Hongbin Pei}, \bibinfo{person}{Bingzhe Wei},
  \bibinfo{person}{Kevin~Chen{-}Chuan Chang}, \bibinfo{person}{Yu Lei}, {and}
  \bibinfo{person}{Bo Yang}.} \bibinfo{year}{2020}\natexlab{}.
\newblock \showarticletitle{Geom-GCN: Geometric Graph Convolutional Networks}.
  In \bibinfo{booktitle}{\emph{8th International Conference on Learning
  Representations, {ICLR} 2020, Addis Ababa, Ethiopia, April 26-30, 2020}}.
\newblock


\bibitem[\protect\citeauthoryear{Rossi, Frasca, Chamberlain, Eynard, Bronstein,
  and Monti}{Rossi et~al\mbox{.}}{2020}]%
        {DBLP:journals/corr/abs-2004-11198}
\bibfield{author}{\bibinfo{person}{Emanuele Rossi}, \bibinfo{person}{Fabrizio
  Frasca}, \bibinfo{person}{Ben Chamberlain}, \bibinfo{person}{Davide Eynard},
  \bibinfo{person}{Michael~M. Bronstein}, {and} \bibinfo{person}{Federico
  Monti}.} \bibinfo{year}{2020}\natexlab{}.
\newblock \showarticletitle{{SIGN:} Scalable Inception Graph Neural Networks}.
\newblock \bibinfo{journal}{\emph{CoRR}}  \bibinfo{volume}{abs/2004.11198}
  (\bibinfo{year}{2020}).
\newblock


\bibitem[\protect\citeauthoryear{Spinelli, Scardapane, and Uncini}{Spinelli
  et~al\mbox{.}}{2020}]%
        {spinelli2020adaptive}
\bibfield{author}{\bibinfo{person}{Indro Spinelli}, \bibinfo{person}{Simone
  Scardapane}, {and} \bibinfo{person}{Aurelio Uncini}.}
  \bibinfo{year}{2020}\natexlab{}.
\newblock \showarticletitle{Adaptive propagation graph convolutional network}.
\newblock \bibinfo{journal}{\emph{IEEE Transactions on Neural Networks and
  Learning Systems}} (\bibinfo{year}{2020}).
\newblock


\bibitem[\protect\citeauthoryear{Tripathy, Yelick, and Buluc}{Tripathy
  et~al\mbox{.}}{2020}]%
        {tripathy2020reducing}
\bibfield{author}{\bibinfo{person}{Alok Tripathy}, \bibinfo{person}{Katherine
  Yelick}, {and} \bibinfo{person}{Aydin Buluc}.}
  \bibinfo{year}{2020}\natexlab{}.
\newblock \showarticletitle{Reducing communication in graph neural network
  training}.
\newblock \bibinfo{journal}{\emph{arXiv preprint arXiv:2005.03300}}
  (\bibinfo{year}{2020}).
\newblock


\bibitem[\protect\citeauthoryear{Velickovic, Cucurull, Casanova, Romero,
  Li{\`{o}}, and Bengio}{Velickovic et~al\mbox{.}}{2018}]%
        {DBLP:conf/iclr/VelickovicCCRLB18}
\bibfield{author}{\bibinfo{person}{Petar Velickovic}, \bibinfo{person}{Guillem
  Cucurull}, \bibinfo{person}{Arantxa Casanova}, \bibinfo{person}{Adriana
  Romero}, \bibinfo{person}{Pietro Li{\`{o}}}, {and} \bibinfo{person}{Yoshua
  Bengio}.} \bibinfo{year}{2018}\natexlab{}.
\newblock \showarticletitle{Graph Attention Networks}. In
  \bibinfo{booktitle}{\emph{6th International Conference on Learning
  Representations, {ICLR} 2018, Vancouver, BC, Canada, April 30 - May 3, 2018,
  Conference Track Proceedings}}. \bibinfo{publisher}{OpenReview.net}.
\newblock


\bibitem[\protect\citeauthoryear{Wang, Ye, and Gupta}{Wang
  et~al\mbox{.}}{2018b}]%
        {wang2018zero}
\bibfield{author}{\bibinfo{person}{Xiaolong Wang}, \bibinfo{person}{Yufei Ye},
  {and} \bibinfo{person}{Abhinav Gupta}.} \bibinfo{year}{2018}\natexlab{b}.
\newblock \showarticletitle{Zero-shot recognition via semantic embeddings and
  knowledge graphs}. In \bibinfo{booktitle}{\emph{Proceedings of the IEEE
  conference on computer vision and pattern recognition}}.
  \bibinfo{pages}{6857--6866}.
\newblock


\bibitem[\protect\citeauthoryear{Wang, Chen, Ren, Yu, Cheng, and Lin}{Wang
  et~al\mbox{.}}{2018a}]%
        {wang2018deep}
\bibfield{author}{\bibinfo{person}{Zhouxia Wang}, \bibinfo{person}{Tianshui
  Chen}, \bibinfo{person}{Jimmy Ren}, \bibinfo{person}{Weihao Yu},
  \bibinfo{person}{Hui Cheng}, {and} \bibinfo{person}{Liang Lin}.}
  \bibinfo{year}{2018}\natexlab{a}.
\newblock \showarticletitle{Deep reasoning with knowledge graph for social
  relationship understanding}.
\newblock \bibinfo{journal}{\emph{arXiv preprint arXiv:1807.00504}}
  (\bibinfo{year}{2018}).
\newblock


\bibitem[\protect\citeauthoryear{Wu, Souza, Zhang, Fifty, Yu, and
  Weinberger}{Wu et~al\mbox{.}}{2019}]%
        {wu2019simplifying}
\bibfield{author}{\bibinfo{person}{Felix Wu}, \bibinfo{person}{Amauri Souza},
  \bibinfo{person}{Tianyi Zhang}, \bibinfo{person}{Christopher Fifty},
  \bibinfo{person}{Tao Yu}, {and} \bibinfo{person}{Kilian Weinberger}.}
  \bibinfo{year}{2019}\natexlab{}.
\newblock \showarticletitle{Simplifying graph convolutional networks}. In
  \bibinfo{booktitle}{\emph{International conference on machine learning}}.
  PMLR, \bibinfo{pages}{6861--6871}.
\newblock


\bibitem[\protect\citeauthoryear{Wu, Pan, Chen, Long, Zhang, and Philip}{Wu
  et~al\mbox{.}}{2020}]%
        {wu2020comprehensive}
\bibfield{author}{\bibinfo{person}{Zonghan Wu}, \bibinfo{person}{Shirui Pan},
  \bibinfo{person}{Fengwen Chen}, \bibinfo{person}{Guodong Long},
  \bibinfo{person}{Chengqi Zhang}, {and} \bibinfo{person}{S~Yu Philip}.}
  \bibinfo{year}{2020}\natexlab{}.
\newblock \showarticletitle{A comprehensive survey on graph neural networks}.
\newblock \bibinfo{journal}{\emph{IEEE transactions on neural networks and
  learning systems}} (\bibinfo{year}{2020}).
\newblock


\bibitem[\protect\citeauthoryear{Xu, Li, Tian, Sonobe, Kawarabayashi, and
  Jegelka}{Xu et~al\mbox{.}}{2018}]%
        {DBLP:conf/icml/XuLTSKJ18}
\bibfield{author}{\bibinfo{person}{Keyulu Xu}, \bibinfo{person}{Chengtao Li},
  \bibinfo{person}{Yonglong Tian}, \bibinfo{person}{Tomohiro Sonobe},
  \bibinfo{person}{Ken{-}ichi Kawarabayashi}, {and} \bibinfo{person}{Stefanie
  Jegelka}.} \bibinfo{year}{2018}\natexlab{}.
\newblock \showarticletitle{Representation Learning on Graphs with Jumping
  Knowledge Networks}. In \bibinfo{booktitle}{\emph{ICML}}.
  \bibinfo{pages}{5449--5458}.
\newblock


\bibitem[\protect\citeauthoryear{Ying, He, Chen, Eksombatchai, Hamilton, and
  Leskovec}{Ying et~al\mbox{.}}{2018}]%
        {ying2018graph}
\bibfield{author}{\bibinfo{person}{Rex Ying}, \bibinfo{person}{Ruining He},
  \bibinfo{person}{Kaifeng Chen}, \bibinfo{person}{Pong Eksombatchai},
  \bibinfo{person}{William~L Hamilton}, {and} \bibinfo{person}{Jure Leskovec}.}
  \bibinfo{year}{2018}\natexlab{}.
\newblock \showarticletitle{Graph convolutional neural networks for web-scale
  recommender systems}. In \bibinfo{booktitle}{\emph{Proceedings of the 24th
  ACM SIGKDD International Conference on Knowledge Discovery \& Data Mining}}.
  \bibinfo{pages}{974--983}.
\newblock


\bibitem[\protect\citeauthoryear{Zeng, Zhou, Srivastava, Kannan, and
  Prasanna}{Zeng et~al\mbox{.}}{2020}]%
        {DBLP:conf/iclr/ZengZSKP20}
\bibfield{author}{\bibinfo{person}{Hanqing Zeng}, \bibinfo{person}{Hongkuan
  Zhou}, \bibinfo{person}{Ajitesh Srivastava}, \bibinfo{person}{Rajgopal
  Kannan}, {and} \bibinfo{person}{Viktor~K. Prasanna}.}
  \bibinfo{year}{2020}\natexlab{}.
\newblock \showarticletitle{GraphSAINT: Graph Sampling Based Inductive Learning
  Method}. In \bibinfo{booktitle}{\emph{8th International Conference on
  Learning Representations, {ICLR} 2020, Addis Ababa, Ethiopia, April 26-30,
  2020}}. \bibinfo{publisher}{OpenReview.net}.
\newblock


\bibitem[\protect\citeauthoryear{Zhang, Huang, Liu, Hu, Song, Ge, Zhang, Wang,
  Zhou, and Qi}{Zhang et~al\mbox{.}}{2020}]%
        {zhang2020agl}
\bibfield{author}{\bibinfo{person}{Dalong Zhang}, \bibinfo{person}{Xin Huang},
  \bibinfo{person}{Ziqi Liu}, \bibinfo{person}{Zhiyang Hu},
  \bibinfo{person}{Xianzheng Song}, \bibinfo{person}{Zhibang Ge},
  \bibinfo{person}{Zhiqiang Zhang}, \bibinfo{person}{Lin Wang},
  \bibinfo{person}{Jun Zhou}, {and} \bibinfo{person}{Yuan Qi}.}
  \bibinfo{year}{2020}\natexlab{}.
\newblock \showarticletitle{AGL: a scalable system for industrial-purpose graph
  machine learning}.
\newblock \bibinfo{journal}{\emph{arXiv preprint arXiv:2003.02454}}
  (\bibinfo{year}{2020}).
\newblock


\bibitem[\protect\citeauthoryear{Zhang and Chen}{Zhang and Chen}{2017}]%
        {zhang2017weisfeiler}
\bibfield{author}{\bibinfo{person}{Muhan Zhang} {and} \bibinfo{person}{Yixin
  Chen}.} \bibinfo{year}{2017}\natexlab{}.
\newblock \showarticletitle{Weisfeiler-lehman neural machine for link
  prediction}. In \bibinfo{booktitle}{\emph{Proceedings of the 23rd ACM SIGKDD
  International Conference on Knowledge Discovery and Data Mining}}.
  \bibinfo{pages}{575--583}.
\newblock


\bibitem[\protect\citeauthoryear{Zhang and Chen}{Zhang and Chen}{2018}]%
        {zhang2018link}
\bibfield{author}{\bibinfo{person}{Muhan Zhang} {and} \bibinfo{person}{Yixin
  Chen}.} \bibinfo{year}{2018}\natexlab{}.
\newblock \showarticletitle{Link prediction based on graph neural networks}.
\newblock \bibinfo{journal}{\emph{arXiv preprint arXiv:1802.09691}}
  (\bibinfo{year}{2018}).
\newblock


\bibitem[\protect\citeauthoryear{Zheng, Ma, Wang, Zhou, Su, Song, Gan, Zhang,
  and Karypis}{Zheng et~al\mbox{.}}{2020}]%
        {zheng2020distdgl}
\bibfield{author}{\bibinfo{person}{Da Zheng}, \bibinfo{person}{Chao Ma},
  \bibinfo{person}{Minjie Wang}, \bibinfo{person}{Jinjing Zhou},
  \bibinfo{person}{Qidong Su}, \bibinfo{person}{Xiang Song},
  \bibinfo{person}{Quan Gan}, \bibinfo{person}{Zheng Zhang}, {and}
  \bibinfo{person}{George Karypis}.} \bibinfo{year}{2020}\natexlab{}.
\newblock \showarticletitle{DistDGL: Distributed Graph Neural Network Training
  for Billion-Scale Graphs}.
\newblock \bibinfo{journal}{\emph{arXiv preprint arXiv:2010.05337}}
  (\bibinfo{year}{2020}).
\newblock


\bibitem[\protect\citeauthoryear{Zhu, Zhao, Yang, Lin, Zhou, Ai, Li, and
  Zhou}{Zhu et~al\mbox{.}}{2019}]%
        {zhu2019aligraph}
\bibfield{author}{\bibinfo{person}{Rong Zhu}, \bibinfo{person}{Kun Zhao},
  \bibinfo{person}{Hongxia Yang}, \bibinfo{person}{Wei Lin},
  \bibinfo{person}{Chang Zhou}, \bibinfo{person}{Baole Ai},
  \bibinfo{person}{Yong Li}, {and} \bibinfo{person}{Jingren Zhou}.}
  \bibinfo{year}{2019}\natexlab{}.
\newblock \showarticletitle{Aligraph: A comprehensive graph neural network
  platform}.
\newblock \bibinfo{journal}{\emph{arXiv preprint arXiv:1902.08730}}
  (\bibinfo{year}{2019}).
\newblock


\end{thebibliography}

\end{document}